\documentclass{article}

 \usepackage[main,final]{neurips_2026}

\usepackage{bbm}
\usepackage[utf8]{inputenc} % allow utf-8 input
\usepackage[T1]{fontenc}    % use 8-bit T1 fonts
\usepackage{hyperref}       % hyperlinks
\usepackage{url}            % simple URL typesetting
\usepackage{booktabs}       % professional-quality tables
\usepackage{amsfonts}       % blackboard math symbols
\usepackage{nicefrac}       % compact symbols for 1/2, etc.
\usepackage{microtype}      % microtypography
\usepackage{xcolor}         % colors
\usepackage{graphicx}      % pictures
\usepackage{todonotes}
\usepackage{bm}
\usepackage{amsmath}
\usepackage{comment}
\usepackage{wrapfig}
\usepackage{enumitem}
\usepackage{xcolor}

\makeatletter
\renewcommand{\paragraph}{%

  \@startsection{paragraph}{4}%
  {\z@}{1ex \@plus .5ex \@minus .2ex}{-1em}%
  {\normalfont\normalsize\bfseries}%
}
\makeatother

\newcommand{\method}{\textcolor{blue}{BLARM}} %< I changed 
\title{\method: Animating 3D Objects from Video via\\ 
\textcolor{blue}{B}lending
\textcolor{blue}{LA}tent 
\textcolor{blue}{R}igid \textcolor{blue}{M}otion Primitives}

\author{
\begin{tabular}{c}
\textbf{Pradyumn Goyal}$^{1,2}$\footnotemark[1] \quad
\textbf{Yizhak Ben-Shabat}$^{1}$ \quad
\textbf{Hsueh-Ti Derek Liu}$^{1}$ \\
\textbf{Haomiao Jiang}$^{1}$ \quad
\textbf{Snehasish Mukherjee}$^{1}$ \quad
\textbf{Kyle Spence}$^{1}$ \\
\textbf{Mark Stauber}$^{1}$ \quad
\textbf{Evangelos Kalogerakis}$^{2,3}$ \quad
\textbf{Yunze Zeng}$^{1}$\footnotemark[2] \\
\\[-0.5em]
$^{1}$Roblox \quad
$^{2}$UMass Amherst \quad
$^{3}$TU Crete
\end{tabular}
}
\renewcommand{\thefootnote}{\fnsymbol{footnote}}

\begin{document}

\maketitle
\footnotetext[1]{Work done primarily during internship at Roblox.}
\footnotetext[2]{Corresponding author.}

% Use numbered footnotes for the paper text.
\renewcommand{\thefootnote}{\arabic{footnote}}
\setcounter{footnote}{0}

\begin{center}
\vspace{-32pt}
    \includegraphics[width=\textwidth]{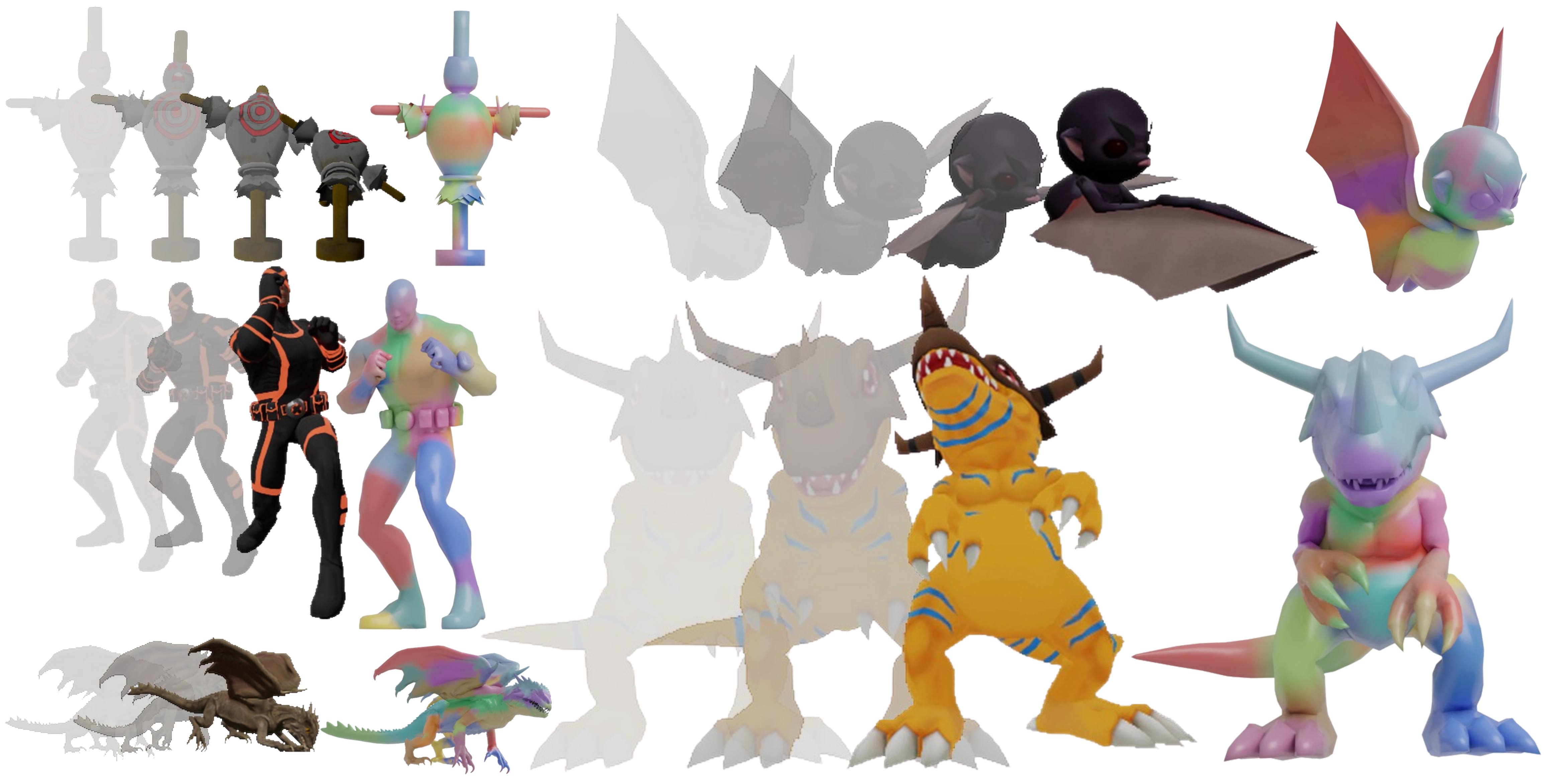}

    \vspace{-0.5em}
    \small
\textbf{\method{}}\footnote{Project page: \url{https://prady1272.github.io/BLARM/}} animates static 3D meshes from monocular videos using compact, time-varying latent rigid motions
and time-invariant vertex-to-component skinning weights, visualized right of the textured animations. Our representation produces temporally coherent, rig-free mesh animations across diverse objects.
    \label{fig:teaser}
\end{center}

%%%%% NEW MATH DEFINITIONS %%%%%

\def\eps{{\epsilon}}

% Vectors
\def\vzero{{\bm{0}}}
\def\vone{{\bm{1}}}
\def\vmu{{\bm{\mu}}}
\def\vtheta{{\bm{\theta}}}
\def\va{{\bm{a}}}
\def\vb{{\bm{b}}}
\def\vc{{\bm{c}}}
\def\vd{{\bm{d}}}
\def\ve{{\bm{e}}}
\def\vf{{\bm{f}}}
\def\vg{{\bm{g}}}
\def\vh{{\bm{h}}}
\def\vi{{\bm{i}}}
\def\vj{{\bm{j}}}
\def\vk{{\bm{k}}}
\def\vl{{\bm{l}}}
\def\vm{{\bm{m}}}
\def\vn{{\bm{n}}}
\def\vo{{\bm{o}}}
\def\vp{{\bm{p}}}
\def\vq{{\bm{q}}}
\def\vr{{\bm{r}}}
\def\vs{{\bm{s}}}
\def\vt{{\bm{t}}}
\def\vu{{\bm{u}}}
\def\vv{{\bm{v}}}
\def\vw{{\bm{w}}}
\def\vx{{\bm{x}}}
\def\vy{{\bm{y}}}
\def\vz{{\bm{z}}}

% Elements of vectors
\def\evalpha{{\alpha}}
\def\evbeta{{\beta}}
\def\evepsilon{{\epsilon}}
\def\evlambda{{\lambda}}
\def\evomega{{\omega}}
\def\evmu{{\mu}}
\def\evpsi{{\psi}}
\def\evsigma{{\sigma}}
\def\evtheta{{\theta}}
\def\eva{{a}}
\def\evb{{b}}
\def\evc{{c}}
\def\evd{{d}}
\def\eve{{e}}
\def\evf{{f}}
\def\evg{{g}}
\def\evh{{h}}
\def\evi{{i}}
\def\evj{{j}}
\def\evk{{k}}
\def\evl{{l}}
\def\evm{{m}}
\def\evn{{n}}
\def\evo{{o}}
\def\evp{{p}}
\def\evq{{q}}
\def\evr{{r}}
\def\evs{{s}}
\def\evt{{t}}
\def\evu{{u}}
\def\evv{{v}}
\def\evw{{w}}
\def\evx{{x}}
\def\evy{{y}}
\def\evz{{z}}

% Matrix
\def\mA{{\bm{A}}}
\def\mB{{\bm{B}}}
\def\mC{{\bm{C}}}
\def\mD{{\bm{D}}}
\def\mE{{\bm{E}}}
\def\mF{{\bm{F}}}
\def\mG{{\bm{G}}}
\def\mH{{\bm{H}}}
\def\mI{{\bm{I}}}
\def\mJ{{\bm{J}}}
\def\mK{{\bm{K}}}
\def\mL{{\bm{L}}}
\def\mM{{\bm{M}}}
\def\mN{{\bm{N}}}
\def\mO{{\bm{O}}}
\def\mP{{\bm{P}}}
\def\mQ{{\bm{Q}}}
\def\mR{{\bm{R}}}
\def\mS{{\bm{S}}}
\def\mT{{\bm{T}}}
\def\mU{{\bm{U}}}
\def\mV{{\bm{V}}}
\def\mW{{\bm{W}}}
\def\mX{{\bm{X}}}
\def\mY{{\bm{Y}}}
\def\mZ{{\bm{Z}}}
\def\mBeta{{\bm{\beta}}}
\def\mPhi{{\bm{\Phi}}}
\def\mLambda{{\bm{\Lambda}}}
\def\mSigma{{\bm{\Sigma}}}

% Entries of a matrix
\def\emLambda{{\Lambda}}
\def\emA{{A}}
\def\emB{{B}}
\def\emC{{C}}
\def\emD{{D}}
\def\emE{{E}}
\def\emF{{F}}
\def\emG{{G}}
\def\emH{{H}}
\def\emI{{I}}
\def\emJ{{J}}
\def\emK{{K}}
\def\emL{{L}}
\def\emM{{M}}
\def\emN{{N}}
\def\emO{{O}}
\def\emP{{P}}
\def\emQ{{Q}}
\def\emR{{R}}
\def\emS{{S}}
\def\emT{{T}}
\def\emU{{U}}
\def\emV{{V}}
\def\emW{{W}}
\def\emX{{X}}
\def\emY{{Y}}
\def\emZ{{Z}}
\def\emSigma{{\Sigma}}

\newcommand{\E}{\mathbb{E}}
\newcommand{\Ls}{\mathcal{L}}
\newcommand{\R}{\mathbb{R}}
\newcommand{\emp}{\tilde{p}}
\newcommand{\lr}{\alpha}
\newcommand{\reg}{\lambda}
\newcommand{\rect}{\mathrm{rectifier}}
\newcommand{\softmax}{\mathrm{softmax}}
\newcommand{\sigmoid}{\sigma}
\newcommand{\softplus}{\zeta}
\newcommand{\KL}{D_{\mathrm{KL}}}
\newcommand{\Var}{\mathrm{Var}}
\newcommand{\standarderror}{\mathrm{SE}}
\newcommand{\Cov}{\mathrm{Cov}}
% Wolfram Mathworld says $L^2$ is for function spaces and $\ell^2$ is for vectors
% But then they seem to use $L^2$ for vectors throughout the site, and so does
% wikipedia.
\newcommand{\normlzero}{L^0}
\newcommand{\normlone}{L^1}
\newcommand{\normltwo}{L^2}
\newcommand{\normlp}{L^p}
\newcommand{\normmax}{L^\infty}

\newcommand{\parents}{Pa} % See usage in notation.tex. Chosen to match Daphne's book.

\begin{abstract}
\vspace{-3pt}

We introduce \method{}, a feed-forward method for video-driven 3D mesh animation. Given a monocular video and a static object mesh, \method{} predicts a temporally coherent animated mesh whose motion follows the video. Rather than relying on explicit rigs or directly regressing high-dimensional vertex motion, we represent animation using a compact set of learned, time-varying rigid motion components and time-invariant vertex-to-component skinning weights. This yields a low-dimensional deformation space without requiring skeletons, cages, skinning weights, or rig annotations. Our architecture conditions geometry-derived deformation latents on video features through factorized spatial-temporal attention, then decodes rigid transformations blended by predicted skinning weights. Trained with trajectory reconstruction, entropy regularization, and motion-aware contrastive learning, \method{} produces accurate and temporally stable animations while recovering compact, interpretable motion structure from monocular video. 
\end{abstract}

\section{Introduction}
\label{sec:intro}
\vspace{-2pt}
Creating dynamic 4D assets is central to graphics, vision, and digital content creation. While recent generative models produce high-quality static 3D objects, animation remains challenging: the representation must preserve geometry and appearance while being temporally coherent and remaining easy to edit and use. Meshes remain a dominant representation in graphics pipelines because they are compact, editable, and broadly supported by tools for rendering, simulation, and animation. Traditional mesh animation systems rely on structured deformation models such as skeletons and cages ~\citep{magnenat1988joint,lewis2000pose,ju2005mean}. These representations provide low-dimensional control over high-resolution geometry: instead of moving every vertex independently, animators control a small set of handles whose motions are blended across the mesh. This subspace view of animation is powerful because many object motions are intrinsically low-dimensional, with large groups of vertices moving coherently as deformable parts. 

Driving these compact representations, however, requires a source of motion. Monocular video is a natural and widely available choice: it captures rich dynamics and is far less laborious to obtain than hand-authored keyframes. One family of recent learning-based methods predicts rigs or deformation structures for 3D assets~\citep{RigNet,UniRig,Magicarticulate}, but these structures are typically inferred primarily from geometry and may not adapt to the target motion observed in a particular video. As a result, they can fail to represent the sequence-specific deformations needed to match the observed motion.
A separate line of work avoids explicit rigging and directly predicts vertex offsets or per-frame deformed meshes from video~\citep{sabathier2026actionmesh,jiang2026mesh4d,chen2026motion}. These approaches can model complex deformations without committing to a fixed kinematic template. However, operating directly in vertex space is high-dimensional and poorly aligned with standard animation workflows. We propose an intermediate representation that is more structured than vertex offsets, but more motion-adaptive than a fixed or geometry-only rig.

We introduce \method{}, a feed-forward framework for animating a  mesh from a monocular video. The core idea is to represent motion using a compact set of learned, time-varying rigid motion components along with time-invariant vertex-to-component skinning weights. Each latent component is decoded into a rigid transformation over time, while each vertex learns how to blend the motion components. This yields a low-dimensional deformation space without requiring predefined skeletons, cages, or rig annotations during training or inference.
%Our architecture first encodes the input mesh into a compact set of geometry-conditioned deformation latents. These latents are then conditioned on video features through factorized spatial and temporal attention, producing motion-aware latents that are decoded into rigid transformations. A skinning head predicts time-invariant vertex-to-component weights over the motion components. 
To encourage meaningful decompositions, we train the model with trajectory reconstruction, entropy regularization for sparse per-vertex skinning, and a motion-aware contrastive objective that encourages vertices with similar motion to share similar skinning weights. Experiments show that \method{} improves geometric accuracy and temporal consistency compared to several animation baselines. We summarize our contributions as follows:

\begin{itemize}[leftmargin=*, itemsep=0.0em, topsep=0.0em]
\vspace{-3pt}
\item We introduce a compact latent deformation representation in which learned time-varying  motion components are blended by time-invariant vertex-to-component skinning weights. This provides a low-dimensional animation space without requiring explicit skeletons, cages, or rig annotations.
\vspace{-2pt}
\item 
We design a feed-forward transformer architecture that combines geometry-conditioned deformation latents with video features using factorized attention, capturing interactions among motion components within each frame and temporal consistency across frames.
\vspace{-2pt}
\item We introduce training losses that favor accurate trajectory reconstruction, sparse per-vertex skinning, and motion-consistent vertex-to-component assignments, yielding coherent mesh animations.

\end{itemize}
\vspace{-3pt}
\section{Related Work}
\vspace{-4pt}
\label{sec:related_work}

\paragraph{4D asset generation.}
4D generation extends 3D content creation to dynamic assets whose geometry, appearance, or motion evolves over time. Recent methods have explored a variety of dynamic representations, including Gaussian Splat models~\citep{ren2023dreamgaussian4d,zeng2024stag4d,ren2024l4gm,gvfd,Dreammesh4d} and dynamic mesh generation methods~\citep{jiang2024animate3d,AnimateAnyMesh,Chen_2025_ICCV,ShapeGen4D}. Gaussian  representations can produce compelling rendered videos, but they are less directly compatible with standard mesh-based animation and simulation, which remain attractive because they preserve explicit surfaces, vertex correspondences, and compatibility with existing graphics pipelines. We therefore focus on video-driven 4D \emph{mesh} animation. We  refer readers to~\citet{miao2025advances} for a  survey of 4D generation.
\vspace{-3pt}
\paragraph{Video-driven mesh animation.}
Recent methods animate meshes from video by predicting vertex displacements or per-frame deformed meshes in a feed-forward manner~\citep{sabathier2026actionmesh,jiang2026mesh4d,chen2026motion}. These methods are  related to our setting because they take a video and a static mesh as input and output an animated mesh without requiring a predefined rig. Their strength is expressivity: by operating directly in vertex space, they can model complex deformations without committing to an explicit kinematic structure. However, direct vertex-space prediction is high-dimensional, difficult to interpret,  and provides limited inductive bias for coherent part motion. In contrast, our method predicts a compact set of rigid transformations and time-invariant vertex-to-component skinning weights. This intermediate representation constrains the deformation space while retaining the feed-forward, video-conditioned nature of these approaches. Other dynamic mesh methods generate a separate 3D shape for each frame, then rely on registration or post-processing to establish temporal correspondences~\citep{Chen_2025_ICCV,ShapeGen4D}. Such pipelines produce  per-frame geometry, but correspondence recovery is expensive and  introduces temporal inconsistencies. Our method instead maintains a fixed mesh and animates it through predicted transformations and skinning weights, preserving vertex correspondences by construction.
\vspace{-3pt}
\paragraph{Rigging.}
Classical animation pipelines use low-dimensional deformation primitives such as skeletons, bones, cages, handles, and skinning weights~\citep{magnenat1988joint,lewis2000pose,skinningcourse2014,ju2005mean,joshi2007harmonic,jacobson2011bounded}. These methods exploit the observation that many mesh motions are locally coherent: groups of vertices move according to shared transformations rather than independent trajectories. Linear Blend Skinning (LBS) is  widely used because it animates high-resolution meshes by blending a small set of rigid transformations, making the resulting representation compact, editable, and compatible with production tools. Learning-based rigging methods infer skeletons, joints, skinning weights, or deformation handles for static 3D assets~\citep{xu2019volskeleton,RigNet,UniRig,Magicarticulate,riganything,makeani,liu2021deepmetahandles,jakab2021keypointdeformer,he2025category}. These methods provide structured and  reusable deformation spaces, but the inferred structure is  driven by geometry or category-level priors alone. Thus it may poorly match to the specific motion observed in a target video, especially for generic objects whose functional articulation depends on both shape and motion. Our method differs by learning a motion-conditioned deformation representation: the latent rigid components and their transformations are inferred from the input video, while the skinning weights remain attached to the mesh. 
Several works also learn skeleton-free deformation representations for pose transfer and dynamic shape modeling from posed 3D meshes, mesh sequences, or vertex trajectories~\citep{liao2022skeleton,chen2023weakly,10076900,10.1007/978-3-031-73220-1_11,zhang2026rigmo}; in contrast, we infer the deformation representation directly from monocular video given a single static target mesh.
Other works also model articulated objects by estimating part decompositions and mechanical joints, such as revolute or prismatic joints~\citep{goyal2025geopard}; see~\citet{liu2025survey} for a survey. These approaches provide interpretable physical structure when the object can be explained as a rigid-body system with explicit joints, but often rely on assumptions about part segmentation, articulation type, or category-specific structure. Our formulation does not require discrete part annotations or a fixed joint model. Instead, it represents motion with a fixed-capacity set of learned rigid components that can approximate diverse articulated or deformable motions through blending.
\vspace{-3pt}
\paragraph{Motion subspaces for dynamic meshes.}
Our work is also related to methods that reduce mesh animation to a lower-dimensional motion subspace. Prior work has parameterized motion using continuous deformation fields~\citep{niemeyer2019occupancy,xie2022neural}, time-varying Jacobian fields~\citep{aigerman2022neural}, spatial deformation handles~\citep{liu2021deepmetahandles,jakab2021keypointdeformer,he2025category}, temporal bases such as splines~\citep{BiMotion}, or subspace transformations predicted over time~\citep{xu2022morig, Magicpony,3dfauna,
Puppeteer,jiang2026mesh4d,chen2026vips}. These methods demonstrate the value of compact motion representations for scalability and temporal regularity. However, many either rely on predefined deformation structures or optimize sequence-specific models.
\method{} combines the advantages of subspace-based animation and video-conditioned feed-forward prediction. It learns a compact set of latent rigid motion components, predicts their time-varying transformations from video, and assigns vertices to those components with time-invariant skinning weights. The representation is adapted to the observed motion; unlike direct vertex-space methods, it remains compact and interpretable; and unlike explicit articulated-object models, it does not require skeletons, part labels, or predefined joint types.

\begin{comment}

\end{comment}

\vspace{-3pt}
\section{Method}
\begin{figure}[!htbp]
    \centering
    \includegraphics[width=0.95\textwidth]{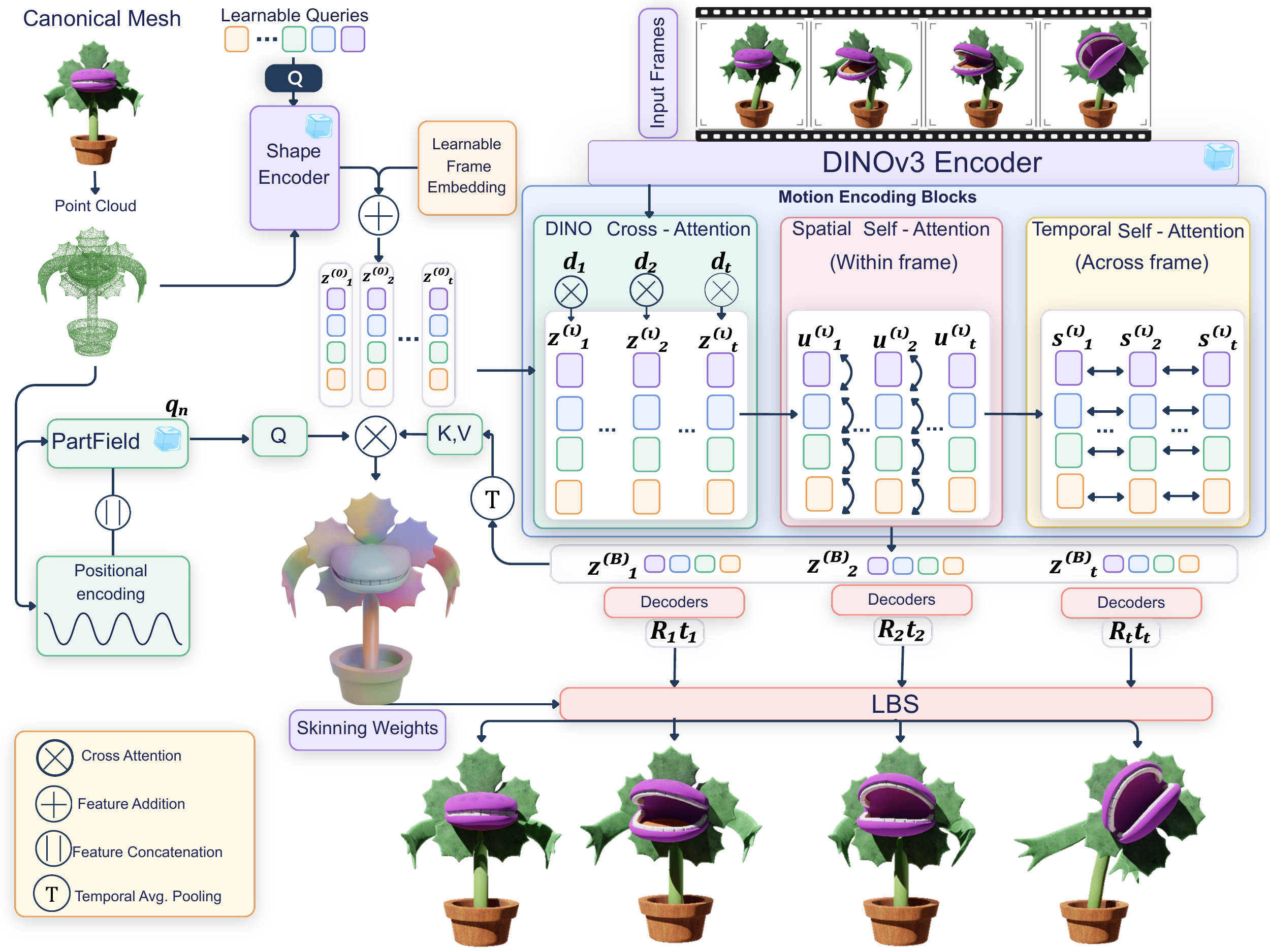}
    \vspace{-3pt}
\caption{
\textbf{Method overview.}
\method{} takes a canonical mesh and a monocular video as input and represents the animation using a compact set of latent rigid motion components. Geometry-aware latents are conditioned on frame-wise DINOv3 features to predict time-varying rigid transformations, while time-invariant vertex-to-component skinning weights are learned \textbf{without explicit skinning supervision}. The final animated sequence is obtained by blending the predicted transformations through linear blend skinning.}
    \label{fig:architecture}
    \vspace{-3pt}
\end{figure}

\vspace{-3pt}
\paragraph{Overview.}
Given a monocular video depicting the motion of an articulated object, the goal of our method is to produce an animated 3D model whose motion is consistent with the video. We assume access to a canonical mesh of the object, optionally with texture, either provided or reconstructed from the first video frame using a state-of-the-art image-to-3D model. In experiments that evaluate animation independently of reconstruction quality, we use the ground-truth canonical mesh when available; otherwise, we use Trellis2~\citep{xiang2025trellis2}. We refer to this mesh as the \emph{canonical mesh}, and denote its vertex positions by $\mX=\{\vx_n\}_{n=1}^{N_v}$, where $N_v$ is the number of vertices. Our method then estimates a temporally coherent animation of this mesh, denoted by $\hat{\mX}_{1:T}=\{\hat{\vx}_{t,n}\}_{t=1,n=1}^{T,N_v}$.
The key idea of our method is to represent object motion using a compact set of time-varying 3D rigid motion components, together with a time-invariant assignment from vertices to those components. This factorization separates \emph{how the object moves} from \emph{which parts of the object are affected by each motion}. We first describe this motion representation, and then the test-time architecture  (Figure \ref{fig:architecture}) used to infer the rigid motion components and vertex-to-component assignments (Section \ref{sec:architecture}). Finally, we describe our training procedure used to learn the model (Section \ref{sec:training}).

\vspace{-3pt}

\vspace{-3pt}
\subsection{Motion Representation}
\vspace{-3pt}
\label{sec:motion_representation}
\paragraph{Latent motion components.}

Classical animation systems often represent shape motion using a sparse set of deformation primitives, such as skeletal joints, bones, cage handles, or sparse simulation particles~\citep{magnenat1988joint,lewis2000pose,joshi2007harmonic,10.1145/1964921.1964987}.

These representations are effective because many object motions are intrinsically low-dimensional: neighboring surface points often follow similar trajectories, and large articulated regions may share a rigid motion induced by a nearby joint or part. Predicting an independent trajectory for every mesh vertex would therefore be highly overparameterized and would make temporal consistency difficult to enforce.

We adopt the same principle, but replace hand-engineered deformation structures with learned latent motion components. This choice is important in our setting because the input is a monocular video and a canonical mesh; we assume no skeletons, cages, or rigging annotations are available for training. Moreover, the objects of interest may not conform to a human-designed kinematic template. Learned latents allow the model to infer a compact object-specific motion basis directly from geometry and video evidence, while still retaining the efficiency and flexibility of primitive-based animation models.

Specifically, for each frame $t$, the model predicts a set of $J$ motion latents
$\{\vz_{t,i}\}_{i=1}^{J}$, each of which is decoded into a rigid transformation $(\mR_{t,i},\vt_{t,i})\in SE(3)$ for $i=1,\dots,J$,
where $\mR_{t,i}\in SO(3)$ and $\vt_{t,i}\in\mathbb{R}^{3}$ denote the rotation and translation associated with the $i$-th latent motion component at frame $t$. These transformations vary over time and define the motion basis used to animate the canonical mesh. We reserve the first component as a special ``root'' component for global object motion, while the remaining components represent local motion relative to the root frame.

\paragraph{Vertex-to-component weights.}
Our representation is inspired by skinning-based animation, where surface points inherit motion from a small number of joints or handles~\citep{magnenat1988joint,lewis2000pose}. For each canonical vertex $\vx_n$, we predict a 
$\vw_n = \{w_{n,i}\}_{i=2}^{J}$
over the non-root latent motion components. The scalar $w_{n,i}$ measures the association between vertex $n$ and latent component $i$, determining how strongly the corresponding rigid transformation contributes to the motion of the vertex. These weights are predicted on the canonical mesh and remain fixed across time. Thus, the temporal variation of the animation is captured by the latent rigid transformations, while the spatial structure of the deformation is captured by the vertex-to-component assignments.
Following standard  properties desired of skinning weights~\citep{lewis2000pose,jacobson2011bounded},
we design the predicted skinning weights to be non-negative, normalized over the non-root components, sparse at the per-vertex level, and spatially coherent. Non-negativity and normalized blending are enforced by a softmax parameterization. Entropy regularization encourages each vertex to depend on a few latent motion components, avoiding dense mixtures of many transformations. Locality and motion-consistent grouping are encouraged by a contrastive learning objective (Section~\ref{sec:training}).

This factorization represents animation through time-varying latent rigid transformations describing \emph{how} the object moves and time-invariant skinning weights describing \emph{where} each component acts.
\vspace{-3pt}
\paragraph{Blended motion.}
Given the latent rigid transformations and vertex-to-component weights, we animate the canonical mesh using linear blend skinning (LBS)~\citep{magnenat1988joint,lewis2000pose}. For each frame $t$ and vertex $\vx_n$, the predicted animated position is
\begin{equation}
\label{eq:lbs_motion_representation}
    \hat{\vx}_{t,n}
    =
    \mR_{t,1}
    \left(
        \sum_{i=2}^{J}
        w_{n,i}
        \left(
            \mR_{t,i}\vx_n + \vt_{t,i}
        \right)
    \right)
    +
    \vt_{t,1}.
\end{equation}
The formulation blends the non-root transformations in the root-local frame and then applies the global root transformation. It is differentiable, and directly compatible with the predicted non-negative normalized weights. It preserves the desired separation between time-varying motion and time-invariant structure: the transformations $(\mR_{t,i},\vt_{t,i})$ change across frames, while the weights $\vw_n$ remain attached to the canonical mesh. As a result, vertices with similar assignments undergo coherent motion over time, even though the latent components themselves are not tied to an explicit skeleton or predefined rig. Note that we use LBS because it provides a minimal and widely used deformation model for blending rigid transformations. Other differentiable skinning formulations, such as dual quaternion skinning~\citep{kavan2007skinning,kavan2008geometric}, could also be used in place of Eq.~\eqref{eq:lbs_motion_representation}. Our contribution is orthogonal to this choice: the central representation is the learned set of latent motion components and the time-invariant vertex-to-component assignment.

\vspace{-3pt}
\subsection{Architecture}
\label{sec:architecture}
\vspace{-3pt}
We now describe the test-time architecture used to infer the motion representation introduced in Section~\ref{sec:motion_representation}. Given a monocular input video and the reconstructed canonical mesh, the network predicts two quantities: a sequence of time-varying latent rigid transformations and a set of time-invariant vertex-to-component skinning weights.
\textbf{Geometry encoding.}
We first encode the canonical mesh into a compact set of geometry-conditioned deformation latents
$\{\vh_i\}_{i=1}^{J}$, where $J$ is much smaller than the number of mesh vertices $N_v$. These latents provide a low-dimensional shape representation from which the model will later infer the motion components. Following query-based shape encoders such as 3DShape2VecSet~\citep{3dshape2vecset}, we uniformly sample $N$ surface points from the input mesh together with their normals. The sampled points are processed by the frozen shape encoder of TripoSG~\citep{li2025triposg}, producing dense geometric features $\{\vg_k\}_{k=1}^{N}$, where $\vg_k \in \mathbb{R}^{D}$.
While these dense features preserve detailed geometric information, they are too numerous to use directly as motion components. We compress them into a smaller set of learned deformation latents.
To this end, we introduce a fixed set of learnable queries
$\{\vq_i\}_{i=1}^{J}$ and let them attend to the dense geometric features through a stack of cross-attention blocks, following DETR's query-based aggregation strategy ~\citep{carion2020detr}:
\begin{equation}
\label{eq:geometry_encoding}
    \{\vh_i\}_{i=1}^{J}
    =
    \mathrm{CrossAttn}_{\mathrm{geom}}
    \left(
        \{\vq_i\}_{i=1}^{J},
        \{\vg_k\}_{k=1}^{N}
    \right).
\end{equation}
The resulting representation $\{\vh_i\}_{i=1}^{J}$ is a compact $J \times D$ set of geometry-aware deformation tokens. Intuitively, each token aggregates information from the canonical shape and acts as a candidate motion component before video conditioning. This compression encourages the subsequent motion encoder to reason over a small set of potential deformation regions rather than over all mesh vertices or sampled surface points.

\textbf{Motion encoding.}
The geometry-conditioned deformation latents $\{\vh_i\}_{i=1}^{J}$ encode the canonical shape, but they are not yet conditioned on the motion observed in the input video. We therefore introduce a stack of motion encoding blocks designed to transform these shape-dependent latents into time-varying motion latents. Each block consists of three operations: cross-attention to per-frame DINOv3 features~\citep{simeoni2025dinov3}, spatial self-attention across deformation latents within each frame, and temporal self-attention along the video sequence.

\textbf{Temporal broadcasting of deformation latents.} We first broadcast the deformation latents across time  
$\vz^{(0)}_{t,i}
    =
    \vh_i,
    t=1,\dots,T,
    i=1,\dots,J$
forming an initial tensor $\mZ^{(0)} \in \mathbb{R}^{T \times J \times D}$. Since the latent index $i$ is shared across frames, this broadcasting step allows each deformation latent  to specialize to a consistent candidate motion region, while its state varies over time according to the video evidence.
To encode frame identity, we add a learnable temporal embedding $\vtheta_t \in \mathbb{R}^{D}$ to all deformation latents at frame $t$:
$\vz^{(0)}_{t,i}
    \leftarrow
    \vz^{(0)}_{t,i}
    +
    \vtheta_t
    =
    \vh_i+\vtheta_t$. The resulting tensor is used as input to our motion encoder.
    
\textbf{Spatiotemporal encoding of visual tokens.} Let $\{\vd_{t,k}\}_{k=1}^{K}$ denote the DINOv3 patch tokens extracted from frame $t$. Because the same local image appearance may occur at different locations or times, we augment these tokens with a spatiotemporal positional encoding. For each patch index $k$ in frame $t$, we compute
$\ve_{t,k}
    =
    f_{\mathrm{pos}}(t,k)
    \in \mathbb{R}^{D},
$
and define $
    \tilde{\vd}_{t,k}
    =
    \vd_{t,k}
    +
    \ve_{t,k}.
$
These position-augmented visual tokens will be used to provide frame-specific evidence to the deformation latents.

\textbf{Motion encoder block.} Each motion encoding block refines the latent tensor through factorized visual, spatial, and temporal attention:
\begin{equation}
\label{eq:motion_encoding_block}
    \mZ^{(\ell)}
    =
    \mathcal{T}^{(\ell)}
    \left(
        \mathcal{S}^{(\ell)}
        \left(
            \mathcal{D}^{(\ell)}
            \left(
                \mZ^{(\ell-1)},
                \tilde{\mD}
            \right)
        \right)
    \right),
    \qquad
    \ell=1,\dots,B,
\end{equation}
where $\mathcal{D}^{(\ell)}$ denotes  cross-attention with DINOv3, $\mathcal{S}^{(\ell)}$ denotes spatial self-attention, and $\mathcal{T}^{(\ell)}$ denotes temporal self-attention, and $\ell$ indexes the motion encoding block. Specifically, within each block, we first condition the deformation latents on the frame-wise visual features to inject frame-specific visual evidence into them. For each frame $t$, the latents attend to the DINOv3 tokens extracted from the same frame:
\begin{equation}
    \{\vu^{(\ell)}_{t,i}\}_{i=1}^{J}
    =
    \mathrm{CrossAttn}_{\mathrm{DINO}}
    \left(
        \{\vz^{(\ell-1)}_{t,i}\}_{i=1}^{J},
        \{\tilde{\vd}_{t,k}\}_{k=1}^{K}
    \right).
\end{equation}

Although each latent maintains a persistent identity over time, object motion often involves dependencies between parts, e.g., the motion of one articulated component may constrain or co-vary with the motion of another. We therefore apply self-attention across the deformation latents  within each frame to reason about interactions among candidate motion components while keeping computation restricted to the compact latent set:
\begin{equation}
    \{\vs^{(\ell)}_{t,i}\}_{i=1}^{J}
    =
    \mathrm{SelfAttn}_{\mathrm{spatial}}
    \left(
        \{\vu^{(\ell)}_{t,i}\}_{i=1}^{J}
    \right).
\end{equation}
\textbf{Temporal self-attention.}
Finally, we model the evolution of each latent component over time. Since the same latent index $i$ corresponds to the same candidate deformation component throughout the sequence, we apply temporal self-attention independently for each latent to aggregate information across frames and encourage temporally coherent motion estimates for each latent component:
\begin{equation}
    \{\vz^{(\ell)}_{t,i}\}_{t=1}^{T}
    =
    \mathrm{SelfAttn}_{\mathrm{temporal}}
    \left(
        \{\vs^{(\ell)}_{t,i}\}_{t=1}^{T}
    \right),
    \qquad
    i=1,\dots,J.
\end{equation}
\textbf{Complexity Analysis.}
Note that compared with full self-attention over all $TJ$ spatiotemporal tokens, which scales as $\mathcal{O}((TJ)^2)$, the factorized spatial--temporal attention used in our motion encoder scales as $\mathcal{O}(T J^2) + \mathcal{O}(J T^2)$.
This makes the motion encoder more efficient while preserving the two interactions needed for animation: coordination among motion components within a frame and temporal consistency of each component across frames. We report run time in Table ~\ref{tab:model_runtime_comparison} in the Appendix.

\textbf{Rigid transformation decoding.}
The output of the motion encoder,
$\mZ^{(B)}=\{\vz^{(B)}_{t,i}\}$, contains one motion-aware latent for each frame $t$ and each latent component $i$. We decode each latent into a rigid transformation that will be used by the skinning model to animate the canonical mesh.
Specifically, each motion latent $\vz^{(B)}_{t,i}$ is passed to two lightweight MLP heads. The first head predicts a 6D continuous rotation representation~\citep{zhou2019continuity}:
\(
    \vr_{t,i}
    = 
    f_{\mathrm{rot}}    
    \left(
        \vz^{(B)}_{t,i}
    \right)
    \in \mathbb{R}^{6}.
\)
We convert this representation into a valid rotation matrix using the Gram--Schmidt orthogonalization procedure of~\citet{zhou2019continuity}:
\(
    \mR_{t,i}
    =
    \mathrm{GS}
    \left(
        \vr_{t,i}
    \right)
    \in SO(3).
\)
The second head predicts the translation component: \(
    \vt_{t,i}
    =
    f_{\mathrm{trans}}
    \left(
        \vz^{(B)}_{t,i}
    \right)
    \in \mathbb{R}^{3}.
\)
Together, $(\mR_{t,i},\vt_{t,i})$ define the rigid motion associated with latent component $i$ at frame $t$. These decoded transformations are subsequently blended using the predicted skinning weights to produce the animated mesh.

\textbf{Skinning weight prediction.}
 For each canonical vertex $\vx_n$, our model predicts skinning weights only over the non-root components, i.e.,
$\vw_n=\{w_{n,i}\}_{i=2}^{J}$, with $w_{n,i}\geq 0$ and $\sum_{i=2}^{J}w_{n,i}=1$.
For each canonical vertex $\vx_n$ with normal $\vn_n$, we first compute a geometric descriptor using a sinusoidal encoding followed by an MLP:
$
    \vp_n
    =
    f_{\mathrm{pt}}
    \left(
        \gamma(\vx_n,\vn_n)
    \right),
$
where $\gamma(\cdot)$ denotes the positional encoding of vertex coordinates and normals.
Since semantic part structure is often correlated with kinematic structure, we additionally use a frozen PartField encoder~\citep{liu2025partfield} to extract a semantic feature $\vs_n$ for each vertex. We note that PartField is used solely to extract frozen semantic features; we do not use any discrete part labels or segmentations derived from its features. We concatenate the geometric and semantic descriptors to obtain $\vq_n = [\vp_n,\vs_n].$
The skinning weights are defined on the canonical mesh and remain fixed, so we summarize each motion component across the video by temporal average pooling,
$\displaystyle \bar{\vz}_i = \frac{1}{T}\sum_{t=1}^{T}\vz^{(B)}_{t,i}$ for $i=1,\dots,J$. 

The vertex descriptors then attend to these time-aggregated latents:
\begin{equation}
    \{\vc_n\}_{n=1}^{N_v}
    =
    \mathrm{CrossAttn}_{\mathrm{skin}}
    \left(
        \{\vq_n\}_{n=1}^{N_v},
        \{\bar{\vz}_i\}_{i=1}^{J}
    \right).
\end{equation}
A lightweight MLP head predicts non-root logits from $\vc_n$, which are normalized with a softmax to obtain the skinning weights $\vw_n$. Note that the root component is excluded from logits regression because it is applied globally to all mesh vertices.

\subsection{Training}
\label{sec:training}
\vspace{-3pt}
We train \method{} using supervision from point trajectories. Given ground-truth animated vertex positions $\vx^{*}_{t,n}$, our objective combines three complementary terms: a trajectory reconstruction loss for accurate animation, an entropy regularizer for sparse per-vertex skinning, and a motion-aware contrastive loss for spatially coherent skinning weights. We ablate entropy and contrastive loss terms in Appendix~\ref{sec:supp_ablation}.
\textbf{Trajectory reconstruction loss.}
The primary training signal encourages the predicted animated mesh to match the observed point trajectories. We measure the reconstruction error between the predicted position $\hat{\vx}_{t,n}$ from Eq.~\eqref{eq:lbs_motion_representation} and the ground-truth position $\vx^{*}_{t,n}$. A uniform point-wise loss, however, can be dominated by static or weakly moving regions. We therefore use a focal-style reweighting~\citep{lin2017focal} that emphasizes points with larger motion error:

\begin{equation}
\label{eq:training_reconstruction_loss}
\begin{aligned}
\mathcal{L}_{\mathrm{rec}}
&=
\frac{1}{T N_v}\sum_{t=1}^{T}\sum_{n=1}^{N_v}
\omega_{t,n}\left\|\hat{\vx}_{t,n}-\vx^{*}_{t,n}\right\|_1,
\qquad
\omega_{t,n}
=
\frac{\left\|\hat{\vx}_{t,n}-\vx^{*}_{t,n}\right\|_2}
{\frac{1}{N_v}\sum_{m=1}^{N_v}\left\|\hat{\vx}_{t,m}-\vx^{*}_{t,m}\right\|_2+\epsilon}.
\end{aligned}
\end{equation}

This weighting encourages the model to explain articulated and deforming regions rather than minimizing the loss primarily through static parts of the mesh.
\textbf{Entropy regularization.}
Reconstruction alone does not prevent a vertex from blending many latent transformations, so we regularize the predicted non-root skinning weights with an entropy penalty to encourage traditional rigging-like behavior, where each vertex is influenced by few bones:
\begin{equation}
\label{eq:training_entropy_loss}
    \mathcal{L}_{\mathrm{ent}}
    =
    -
    \frac{1}{N_v}
    \sum_{n=1}^{N_v}
    \sum_{i=2}^{J}
    w_{n,i}
    \log(w_{n,i}+\epsilon).
\end{equation}

\paragraph{Motion-aware contrastive loss.}
While entropy promotes sparse per-vertex assignments, it does not by itself enforce spatial or kinematic coherence. We therefore add a contrastive loss that encourages vertices with similar motion to have similar skinning weights, and vertices with dissimilar motion to have distinct weights. We construct positive and negative pairs from ground-truth point trajectories. For each vertex, we compute a compact trajectory descriptor $\vm_{t,n}$ encoding translational direction, rotational direction, motion magnitude, and a static-point indicator. We then define motion similarity between vertices by averaging cosine similarity over time:
$
    s^{\mathrm{mot}}_{n,m}
    =
    \frac{1}{T}
    \sum_{t=1}^{T}
    \left\langle
        \vm_{t,n},
        \vm_{t,m}
    \right\rangle .
$
For an anchor vertex $a$, positives are sampled from vertices with high motion similarity, while negatives are sampled from vertices with low motion similarity. We additionally bias positive sampling toward nearby canonical vertices and include both spatially distant easy negatives and nearby hard negatives. Details of the trajectory descriptor and sampling procedure are provided in Appendix~\ref{sec:supp_contrastive_loss}. The contrastive loss is applied to the predicted non-root skinning weights. Given two vertices $u$ and $v$, we measure similarity between their skinning weights using the Bhattacharyya coefficient:
\(
    s^{\mathrm{skin}}_{u,v}
    =
    \sum_{i=2}^{J}
    \sqrt{
        w_{u,i}w_{v,i}
    } .
\)
For an anchor $a$, a positive $b$, and sampled negatives $\mathcal{N}(a)$, we use an InfoNCE-style objective:
\begin{equation}
    \mathcal{L}_{\mathrm{con}}\!\!
    =\!\!
    -\frac{1}{2}\!\!
    \left[
    \log\!\!
    \frac{
        \exp(s^{\mathrm{skin}}_{a,b}/\tau)
    }{
        \exp(s^{\mathrm{skin}}_{a,b}/\tau)
        \!+\!
        \sum_{c\in\mathcal{N}(a)}
        \exp(s^{\mathrm{skin}}_{a,c}/\tau)
    }
    \!+\!
    \log\!\!
    \frac{
        \exp(s^{\mathrm{skin}}_{b,a}/\tau)
    }{
        \exp(s^{\mathrm{skin}}_{b,a}/\tau)
        \!+\!
        \sum_{c\in\mathcal{N}(b)}
        \exp(s^{\mathrm{skin}}_{b,c}/\tau)
    }
    \right]\!\!.
\end{equation}Here, $\tau$ is a temperature parameter. This objective encourages motion-consistent vertices to share similar vertex-to-component assignments while separating vertices that follow different motions.
\textbf{Final objective.}
The full training objective is
\begin{equation}
\label{eq:training_full_objective}
    \mathcal{L}
    =
    \lambda_{\mathrm{rec}}
    \mathcal{L}_{\mathrm{rec}}
    +
    \lambda_{\mathrm{ent}}
    \mathcal{L}_{\mathrm{ent}}
    +
    \lambda_{\mathrm{con}}
    \mathcal{L}_{\mathrm{con}} .
\end{equation}
In our implementation, we set $\lambda_{\mathrm{rec}}=3$, $\lambda_{\mathrm{ent}}=0.001$, and $\lambda_{\mathrm{con}}=0.3$.
%\section{Inference} \pradyumn{Need to add inference section}
\textbf{Implementation details.} More details on the architecture and training are provided in Appendix~\ref{sec:supp_implementation_details}.

\vspace{-3pt}
\section{Experiments}
\label{sec:experiments}
\label{sec:results}
\vspace{-3pt}
We now discuss our training and evaluation protocol, along with comparisons, results, and ablations.
\vspace{-3pt}
\paragraph{Training dataset.}
We train on animated object sequences from the Objaverse-1.0 split curated by~\citet{Diffusion4d}. We remove all objects and sequences that appear in our evaluation sets, resulting in approximately $10k$ shapes. For each sequence, we compute a canonical normalization transform from the first frame by scaling the object to fit inside a unit bounding box, and apply the same transform to all subsequent frames to preserve temporal consistency. We render each normalized sequence as a monocular video at resolution \(512 \times 512\), with uniform sampling of camera view points. The resulting canonical and animated mesh sequences, and rendered videos form our training data.
\textbf{Test datasets.}
We first evaluate on ActionBench~\citep{sabathier2026actionmesh} and Motion80~\citep{chen2026motion}, two benchmarks with textured canonical meshes and ground-truth mesh animations. ActionBench contains 128 16-frame sequences, while Motion80 contains 80 variable-length sequences. On both datasets, we evaluate geometric accuracy directly on the predicted mesh motion and appearance quality using rendered videos. Additional dataset details are provided in the Appendix \ref{sec:supp_eval_datasets}.
We also report qualitative and quantitative results on the in-the-wild split of Consistent4D~\citep{jiang2023consistent4d}, where canonical meshes are not provided. In this setting, we reconstruct the canonical textured mesh using an off-the-shelf image-to-3D model~\citep{xiang2025trellis2}, and then apply our animation model to the reconstructed mesh.
\paragraph{Baselines.}
We compare against three feed-forward mesh animation baselines: ActionMesh~\citep{sabathier2026actionmesh}, Mesh4D~\citep{jiang2026mesh4d}, and Motion 3-to-4~\citep{chen2026motion}. These are the most direct comparisons for our setting because they take a video and a canonical mesh as input and predict an animated mesh without requiring a predefined rig. Unlike these methods, which directly regress vertex displacements or deformed vertex positions, \method{} predicts a compact set of root-relative rigid transformations and blends them with time-invariant skinning weights. This representation constrains the deformation space while remaining feed-forward and rig-free.
\textbf{Evaluation metrics.}
For geometric evaluation on ActionBench~\citep{sabathier2026actionmesh} and Motion80~\citep{chen2026motion}, we follow the ActionMesh protocol and report three complementary metrics \textbf{CD-3D}, \textbf{CD-4D}, \textbf{CD-Motion}; we refer readers to ~\citet{sabathier2026actionmesh} for implementation details. We also evaluate rendered appearance. We report \textbf{LPIPS}, \textbf{CLIP} similarity, \textbf{FVD}, and \textbf{DreamSim}, capturing complementary aspects of perceptual fidelity, semantic consistency, and temporal realism. We refer readers to Appendix \ref{sec:supp_metrics} for additional details.

\begin{figure}[t]
    \centering
    \includegraphics[
        width=1.000\textwidth,
        height=1.000\textheight,
        keepaspectratio
    ]{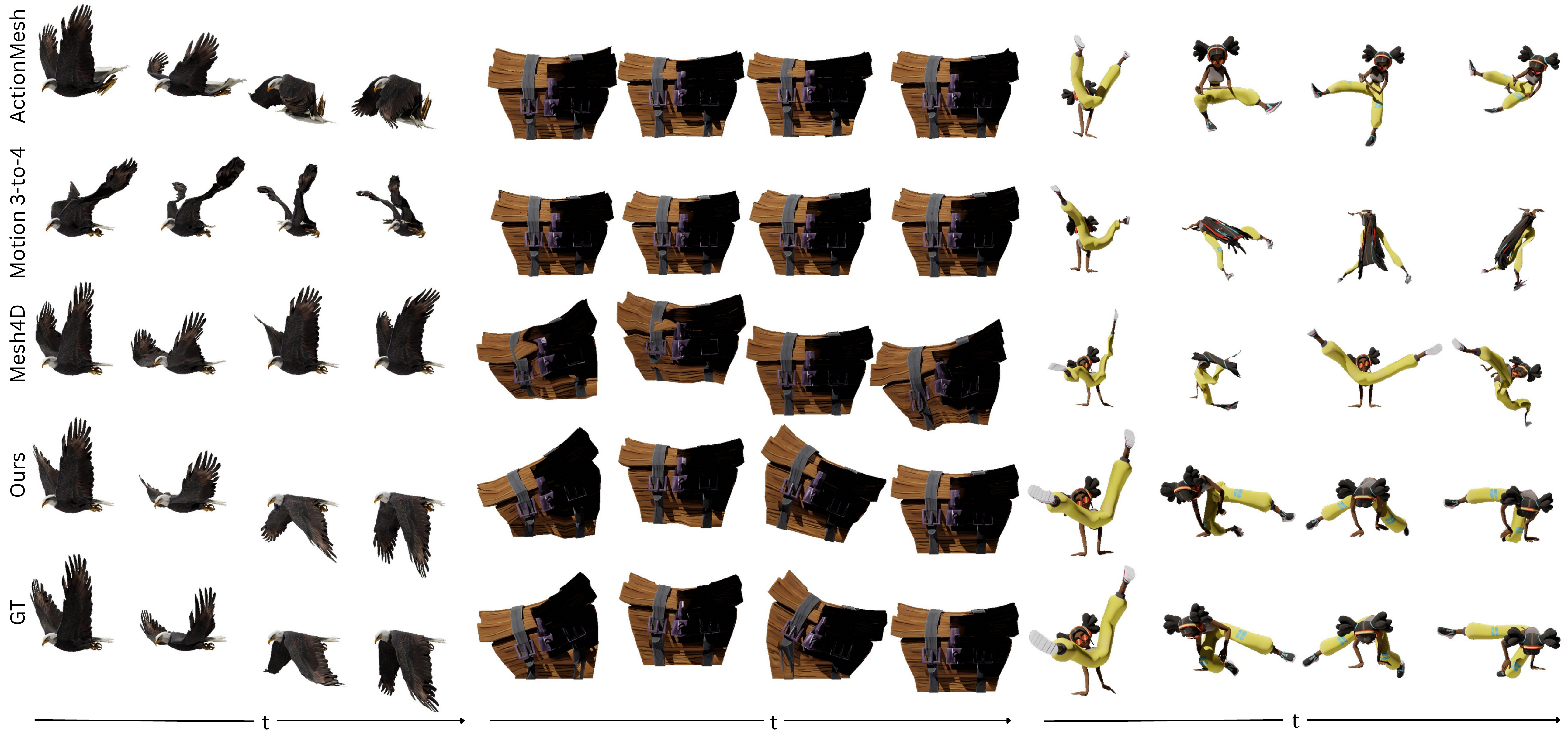}
    \vspace{-20pt}
    \caption{\textbf{Qualitative comparison on ActionBench.}
Across diverse shapes, \method{} preserves the canonical mesh structure better than baselines and yields temporally more coherent animations with fewer jittering or distorted parts. 
% \textbf{We provide MP4s on an HTML page in the supplement.}
} 
    \label{fig:Actionbench_qualitative}
\end{figure}

\begin{table}[t]
\centering
\small
\begin{tabular}{l|ccc|cccc}
\toprule
& \multicolumn{3}{c|}{\textbf{Geometry}}
& \multicolumn{4}{c}{\textbf{Appearance}} \\
\cline{2-8}
\noalign{\vskip 2pt}
\textbf{Method}
& CD-3D$\downarrow$
& CD-4D$\downarrow$
& CD-Motion$\downarrow$
& LPIPS$\downarrow$
& CLIP$\uparrow$
& FVD$\downarrow$
& DreamSim$\downarrow$ \\
\midrule
Mesh4D        & 2.57 & 5.01 & 9.24  & 0.06 & 0.95 & 787.38  & 0.08 \\
Motion 3-to-4  & 2.49 & 5.34 & 10.31 & 0.07 & 0.92 & 1270.80 & 0.12 \\
ActionMesh    & 3.30 & 5.61 & 11.12 & 0.08 & 0.93 & 1126.90 & 0.11 \\
Ours          & \textbf{1.71} & \textbf{3.07} & \textbf{7.15} & \textbf{0.05} & \textbf{0.96} & \textbf{426.72} & \textbf{0.05} \\
\bottomrule
\end{tabular}
\vspace{3pt}
\caption{
\textbf{Quantitative evaluation on ActionBench.}
\method{} outperforms feed-forward mesh animation baselines across geometric and appearance based metrics, indicating more accurate and temporally stable mesh animation.
}
\label{tab:ActionBench}
\end{table}

\begin{table}[t]
\centering
\small
\begin{tabular}{l|ccc|cccc}
\toprule
& \multicolumn{3}{c|}{\textbf{Geometry}}
& \multicolumn{4}{c}{\textbf{Appearance}} \\
\cline{2-8}
\noalign{\vskip 2pt}
\textbf{Method}
& CD-3D$\downarrow$
& CD-4D$\downarrow$
& CD-Motion$\downarrow$
& LPIPS$\downarrow$
& CLIP$\uparrow$
& FVD$\downarrow$
& DreamSim$\downarrow$ \\
\midrule
Mesh4D        & 5.40 & 11.13 & 24.26 & 0.11 & 0.93 & 1517.45 & 0.11 \\
Motion 3-to-4  & 2.25 & 5.89  & 13.00 & 0.07 & 0.95 & 735.57  & 0.06 \\
ActionMesh    & 2.44 & 5.48  & 12.78 & 0.08 & 0.94 & 791.80  & 0.08 \\
Ours          & \textbf{1.93} & \textbf{3.46} & \textbf{8.72} & \textbf{0.06} & \textbf{0.97} & \textbf{482.50} & \textbf{0.05} \\
\bottomrule
\end{tabular}
\vspace{3pt}
\caption{
\textbf{Quantitative evaluation on Motion80.}
\method{} achieves better performance across metrics, demonstrating accurate reconstruction and temporally consistent mesh motion on long animation sequences.
}
\label{tab:Motion80}
\vspace{-10pt}
\end{table}

\paragraph{\textbf{Quantitative results.}}
Tables~\ref{tab:ActionBench} and~\ref{tab:Motion80} report quantitative results on ActionBench and Motion80. We report results for Consistent4D in Appendix~\ref{sec:supp_Results}.
\method{} achieves the best performance on average on all three benchmarks. Lower CD-3D and CD-4D show that our predictions better match the target mesh sequence under frame level alignment, while the larger CD-M improvement reflects more stable, temporally consistent mesh motion.
\method{} also improves appearance-based metrics, achieving the best LPIPS, CLIP similarity, FVD, and DreamSim scores on both datasets. The gain in FVD is particularly important because it is sensitive to temporal artifacts in rendered videos. These results suggest that the proposed root-relative rigid components and time-invariant skinning weights improve not only geometric alignment, but also the visual stability of rendered animations.
\begin{figure}[t]
    \centering
    \vspace{-3pt}
    \includegraphics[width=\textwidth]{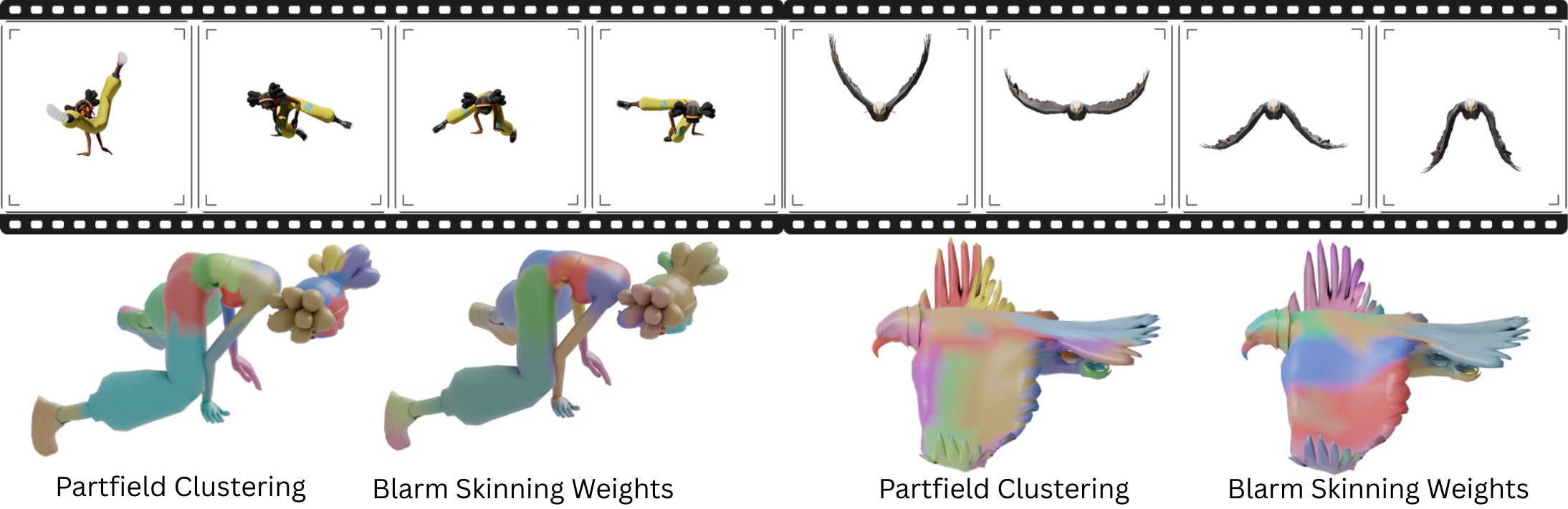}
    \vspace{-3pt}
    \caption{
    \textbf{PartField clustering vs. BLARM skinning weights.} 
    We compare clusters of PartField features with \method{}’s learned skinning decomposition. PartField emphasizes semantic part structure, while BLARM groups regions according to shared motion, yielding a distinct motion-aware decomposition.
    }
    \label{fig:skinning_weights}
\end{figure}
\textbf{Qualitative results.} Figure~\ref{fig:Actionbench_qualitative} shows qualitative comparisons on ActionBench. We refer readers to Appendix~\ref{sec:supp_Results} for qualitative results on Motion80. Baseline methods often produce plausible shapes in individual frames, but can exhibit temporal jitter, part distortion, or inconsistent deformation over time. In contrast, \method{} better preserves the structure of the canonical mesh while producing stable motion across the sequence.

\paragraph{\textbf{Skinning decomposition.}}
Figure~\ref{fig:skinning_weights} compares the predicted skinning weights with a $k{=}31$ clustering of the PartField features used by our model. Each latent component or feature cluster is assigned a distinct color. While PartField primarily captures semantic part structure, \method{} learns vertex-to-component assignments that reflect the motion observed in the input video. The resulting maps often align with coherent moving regions, such as bending shoe parts and feather groups. Although these components are learned without explicit rig supervision, the visualizations suggest that the model often discovers structured, part-aware decompositions that support temporally coherent mesh animation.

\vspace{-3pt}
\section{Discussion and Limitations}
\vspace{-3pt}

Our results suggest that representing vertex motion through a compact subspace of learned rigid transformations provides an effective structure for generating geometrically plausible and temporally coherent animations. 
\textbf{Limitations.} First, the learned skinning weights may be suboptimal in challenging cases, causing nearby or visually similar regions to receive incorrect vertex-to-component assignments resulting in less coherent animation. Second, our approach assumes that the input canonical mesh has a topology that can reasonably support the target motion. When this assumption is violated, the predicted deformation may produce artifacts. Our method also requires ground-truth point trajectories for training and does not explicitly model the deformation of interior structures. We also refer readers to Appendix~\ref{sec:supp_Limitations}.

\section*{Acknowledgements}
This project has received funding from the European Research Council (ERC) under the Horizon Research and 
Innovation Programme (Grant agreement No. 101124742).

\bibliographystyle{plainnat}
\bibliography{neurips_2026}
\appendix
\clearpage
\setcounter{page}{1}

\begin{center}
{\Large \bf Appendix}
\end{center}
\appendix
% \section{Appendix}
In the appendix, we provide additional details and experimental results:

\begin{itemize}
    \item Appendix~A presents ablation studies that analyze the contribution of each component.
    \item Appendix~B provides experimental details, including evaluation metrics, datasets, additional results, and limitations.
    \item Appendix~C describes implementation details, including preprocessing, architecture, decoders, training, and inference.
    \item Appendix~D details the motion-aware contrastive loss.
    \item Appendix~E discusses the broader societal impact of the proposed method.
\end{itemize}

\section{Ablations}
\label{sec:supp_ablation}
Table~\ref{tab:ablations} analyzes the contribution of the main components of \method{} on ActionBench. Lat./Vtx. denotes the average number of active non-root components assigned to each vertex, i.e. vertex-to-latent weights above 0.01,  and Lat./Shape denotes the average number of active non-root components used per shape.
Removing PartField features, entropy regularization, and contrastive learning substantially degrades performance, increasing CD-3D, CD-4D, and CD-M to $2.95$, $5.78$, and $13.53$, respectively. In this setting, the model effectively collapses to using very few non-root latent components, as reflected by the latent-count statistics. Removing only the contrastive loss causes a similar collapse, indicating that motion-aware contrastive supervision is critical for encouraging different latent components to specialize to different moving regions. We further compare against a direct per-vertex offset prediction baseline that removes our latent rigid-motion formulation. While it improves over collapsed latent variants, it remains notably worse than the full model, particularly on CD-M, supporting the benefit of structured latent motion for temporally coherent deformation.

Adding PartField features improves motion fidelity, reducing CD-M from $12.90$ to $10.51$, suggesting that semantic vertex representations help align the learned components with meaningful object parts. Entropy regularization further improves performance by encouraging sparse and confident vertex-to-component assignments, reducing CD-M to 9.48 in the controlled setting. Finally, the full model achieves the best reconstruction and motion accuracy, with CD-3D, CD-4D, and CD-M of 1.71, 3.07, and 7.15. These results show that semantic vertex features, sparse skinning weights, and motion-aware contrastive learning are complementary: together they prevent latent collapse, encourage part-level decomposition, and improve temporally coherent mesh animation.

\paragraph{Number of motion primitives.}
We further analyze the effect of the number of non-root latent motion primitives $J$ in Table~\ref{tab:ablation-joints}. 
With $J=15$, the limited deformation capacity leads to worse reconstruction and denser vertex-to-latent assignments. 
Increasing to $J=63$ gives comparable performance to $J=31$, but uses substantially more active latents per shape, while $J=127$ further increases fragmentation and degrades performance. 
Overall, $J=31$ provides a good balance between deformation capacity, sparsity, and representation compactness.

\begin{table*}[t]
\centering
\small
\setlength{\tabcolsep}{5pt}
\begin{tabular}{lccccc}
\toprule
\textbf{Model} &
\textbf{CD-3D}$\downarrow$ &
\textbf{CD-4D}$\downarrow$ &
\textbf{CD-M}$\downarrow$ &
\textbf{Lat./Vtx.} &
\textbf{Lat./Shape} \\
\midrule
w/o PartField, ent., con. & 2.95 & 5.78 & 13.53 & 1.00 & 1.00  \\
w/o contrastive           & 2.95 & 5.88 & 12.90 & 1.00 & 1.00  \\
Per-vertex direct prediction  & 2.47 & 4.33 & 10.96 & ---  & ---   \\
w/o PartField             & 2.37 & 4.07 & 10.51 & 1.85 & 22.20 \\
w/o entropy               & 2.22 & 4.05 & 9.71  & 1.97 & 27.98 \\
Control                   & 2.22 & 4.04 & 9.48  & 1.59 & 25.52 \\
Final model               & \textbf{1.71} & \textbf{3.07} & \textbf{7.15} & 1.71 & 29.50 \\
\bottomrule
\end{tabular}
\caption{
\textbf{Ablation study on ActionBench.}
Each component of \method{} contributes to performance: PartField features, entropy regularization, and motion-aware contrastive learning improve motion accuracy and sparsity while encouraging compact latent usage.
}
\label{tab:ablations}
\end{table*}

\begin{table}[t]
    \centering
    
    \label{tab:num_motion_primitives}
    \begin{tabular}{lccccc}
        \toprule
        Model & CD-3D $\downarrow$ & CD-4D $\downarrow$ & CD-M $\downarrow$ & Lat./Vtx. & Lat./Shape \\
        \midrule
        $J=15$        & 2.53 & 4.49 & 11.42 & 2.63 & 14.64 \\
        $J=31$        & 2.22 & 4.04 & 9.48  & 1.59 & 25.52 \\
        $J=63$        & 2.38 & 4.20 & \textbf{9.44} & \textbf{1.22} & 42.01 \\
        $J=127$       & 2.44 & 5.11 & 11.88 & 1.24 & 96.27 \\
        \bottomrule

    \end{tabular}
  \caption{\textbf{Number of latent motion primitives.} Too few primitives restrict deformation capacity, while too many lead to less compact representations without clear performance gains; $J=31$ provides the best overall balance.}
    \label{tab:ablation-joints}
\end{table}

\begin{figure}[t]
    \centering
    \includegraphics[
        width=1.000\textwidth,
        height=1.000\textheight,
        keepaspectratio
    ]{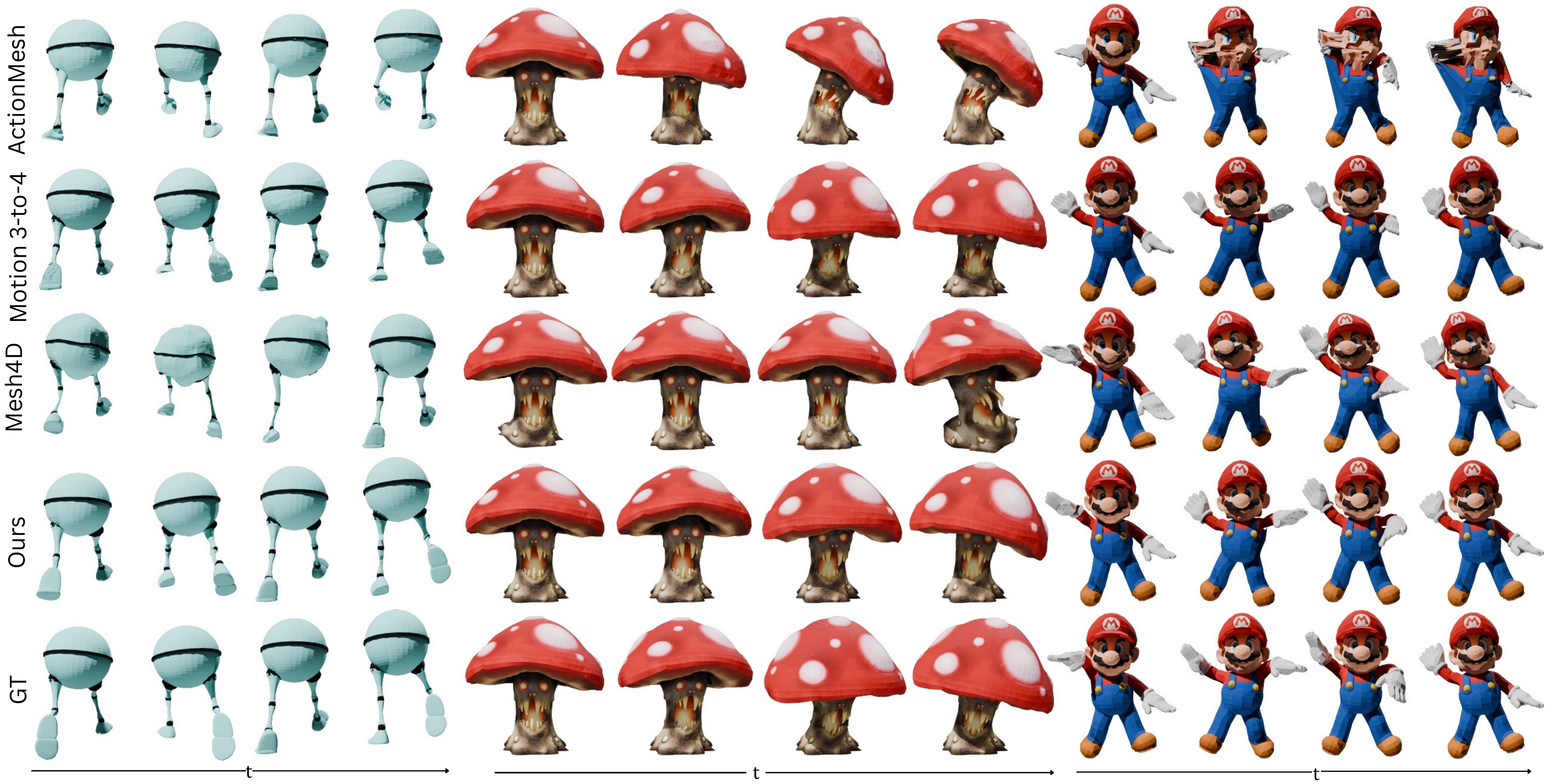}
    \vspace{-20pt}
    \caption{\textbf{Qualitative comparison on Motion80.} Compared to the baselines, \method{} better maintains part structure and produces more stable deformations across time, yielding animations that closely match the ground-truth motion.}
    \label{fig:Motion80_qualitative}
\end{figure}

\begin{figure}[t]
    \centering
    \includegraphics[
        width=1.000\textwidth,
        height=1.000\textheight,
        keepaspectratio
    ]{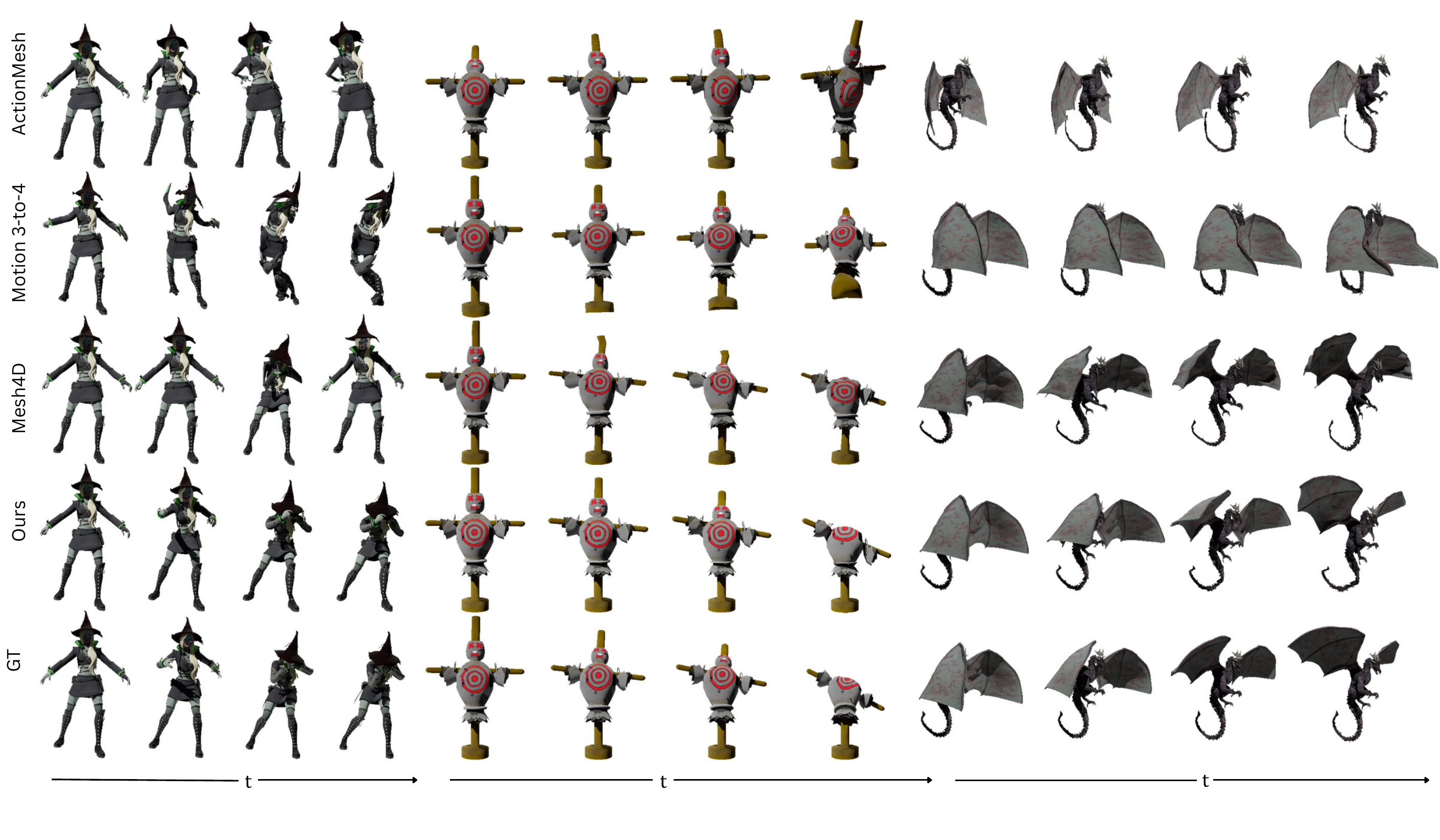}
   \caption{
\textbf{Additional qualitative results on ActionBench.}
We show additional comparisons with feed-forward mesh animation baselines.
}
    \label{fig:qualitative_comparison_supplementary}
\end{figure}

\section{Experiments}
\subsection{Metrics}
\label{sec:supp_metrics}
\paragraph{Geometric Metrics} We here explain how each Geometric metric is calculated and refer readers to ~\citet{sabathier2026actionmesh} for implementation details.
\textbf{CD-3D} measures per-frame reconstruction accuracy by aligning each predicted mesh to the corresponding ground-truth frame with ICP and computing Chamfer Distance. \textbf{CD-4D} evaluates sequence-level consistency by estimating a single global rigid alignment from the first frame and averaging Chamfer Distance over the full sequence. \textbf{CD-Motion} measures motion fidelity by evaluating whether point correspondences are preserved over time: after global ICP alignment, nearest-neighbor correspondences are computed in the first frame and the bidirectional correspondence error is accumulated throughout the sequence. All geometric metrics are computed using \(100{,}000\) uniformly sampled surface points, with meshes centered at the origin and normalized to a unit bounding box.

\paragraph{Appearance Metrics}
 For each predicted textured mesh sequence, we render four evenly spaced viewpoints around the object. When ground-truth textured mesh sequences are available, for ActionBench and Motion80, we render them from the same viewpoints and compare predicted and reference videos.

\subsection{Evaluation Datasets}
\label{sec:supp_eval_datasets}

We provide additional details on the evaluation datasets used in Sec.~\ref{sec:experiments}.

\paragraph{ActionBench.}
ActionBench~\citep{sabathier2026actionmesh} provides object identifiers for animated assets, and not the textured mesh. Since these identifiers are publicly available, we retrieve the corresponding object glbs from Objaverse~\citep{Objaverse}. We render each animation for 16 frames, matching the frame count used by ActionBench. For all methods, we use the ground-truth mesh from the first frame as the canonical input mesh. Since all evaluated baselines predict vertex offsets on the input mesh, we preserve the original textures and use them when rendering predicted animations for appearance metrics and qualitative comparisons.

\paragraph{Motion80.}
Motion80~\citep{chen2026motion} provides animated meshes together with their textures. As in ActionBench, the first-frame mesh is used as the canonical input for all methods, and the provided texture is retained for rendering both predicted and ground-truth animated sequences.

\paragraph{Consistent4D.}
We additionally evaluate on the in-the-wild split of Consistent4D~\citep{jiang2023consistent4d}, which contains 12 video sequences. Since this dataset does not provide ground-truth meshes, we reconstruct the first-frame shape and material using TRELLIS2~\citep{xiang2025trellis2} and provide the same reconstructed canonical mesh to all baselines. 

Since ground-truth meshes and camera parameters are not available, we render the reconstructed animated mesh from four uniformly spaced viewpoints and report appearance metrics against the input video after averaging over the rendered views.

\subsection{Additional Results}
\label{sec:supp_Results}
\begin{figure}[t]
    \centering
    \includegraphics[
        width=1.000\textwidth,
        height=1.000\textheight,
        keepaspectratio
    ]{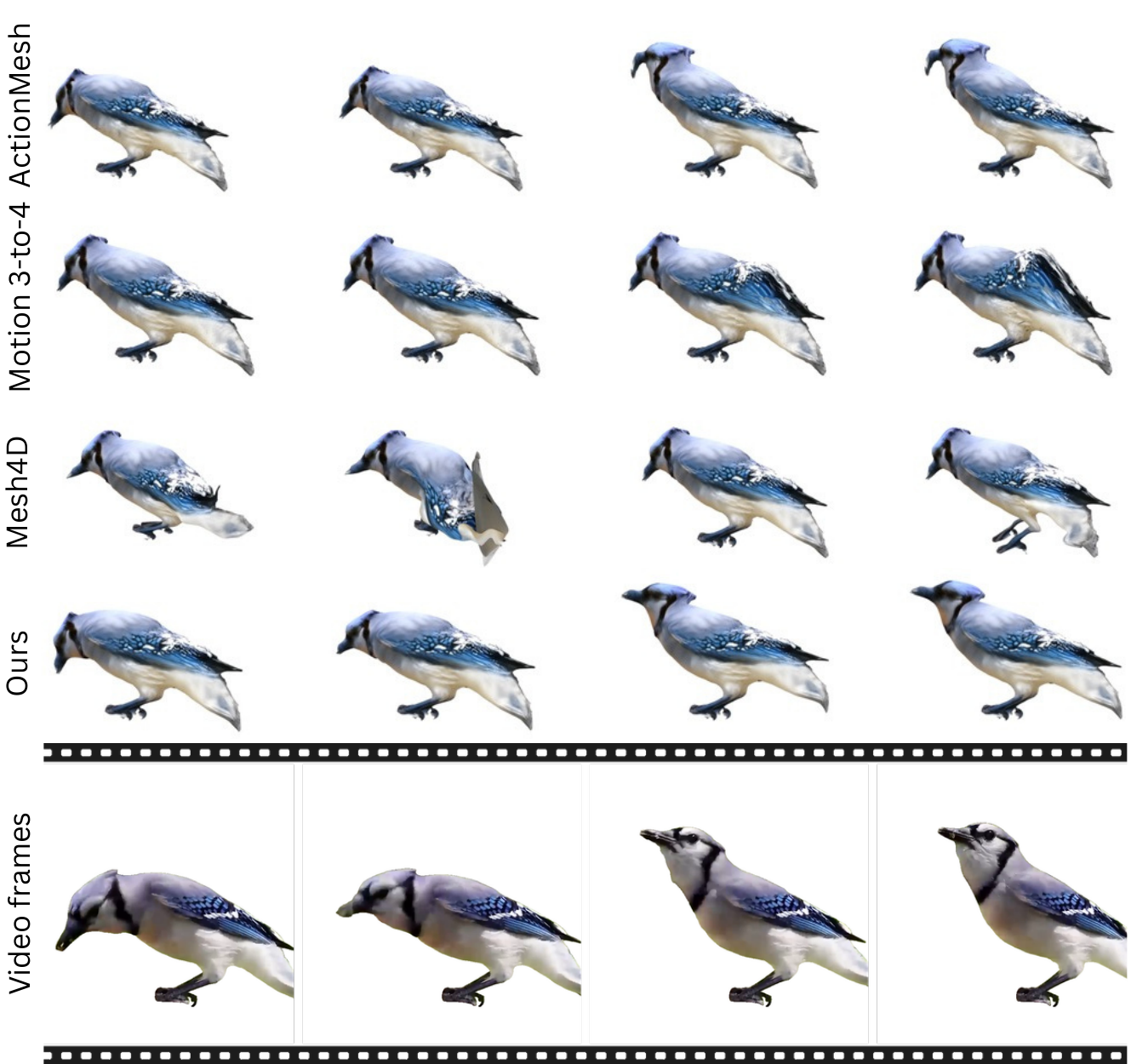}
        \caption{\textbf{Qualitative comparison on Consistent4D.} With reconstructed meshes, \method{} yields stable animations that better match the input video.}
    \label{fig:consistent4D_qualitative}
\end{figure}

Figure \ref{fig:qualitative_comparison_supplementary} and Figure \ref{fig:Motion80_qualitative} show qualitative results on ActionBench, and Motion80, respectively. \method{} produces more temporally stable animations while better preserving the geometrical structure. Compared to baselines, it reduces visible distortions and part entanglement, leading to motion that more closely follows the ground-truth sequence.
On Consistent4D, our method obtains favorable LPIPS, FVD, and DreamSim scores while performing comparably on CLIP in Table~\ref{tab:consistent4d_quant}. The qualitative results in Figure~\ref{fig:consistent4D_qualitative} suggest that our animation maintains the reconstructed canonical structure over time.

\subsection{Limitations}
\label{sec:supp_Limitations}
\begin{figure}[t]
    \centering
    \includegraphics[
        width=0.5000\textwidth,
        height=0.5000\textheight,
        keepaspectratio
    ]{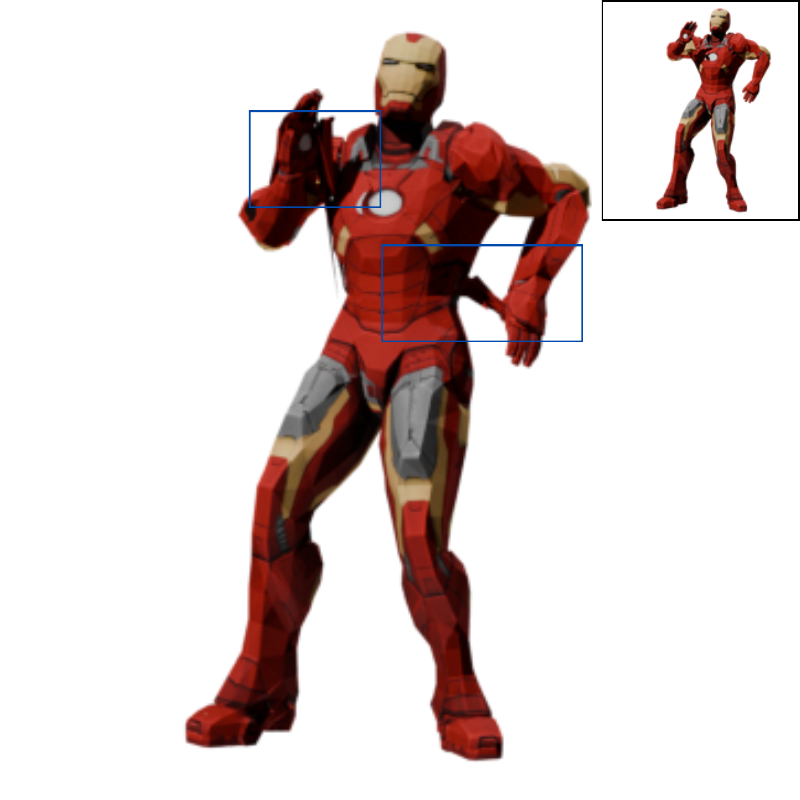}
        \caption{
\textbf{Limitations.}
Incorrect vertex-to-component assignments in visually similar regions can entangle nearby parts, producing less coherent motion.
}
    \label{fig:Limitations}
\end{figure}

Figure \ref{fig:Limitations} showcases our limitation. Our method can produce imperfect mesh animation, when nearby regions have similar visual appearance but should move independently. In such cases, incorrect vertex-to-component assignments may entangle adjacent parts, leading to suboptimal animation.

\begin{table}[t]
\centering
\begin{tabular}{lcccc}
\toprule
\textbf{Method} & \textbf{LPIPS$\downarrow$} & \textbf{CLIP$\uparrow$} & \textbf{FVD$\downarrow$} & \textbf{DreamSim$\downarrow$} \\
\midrule
Mesh4D      & 0.16 & 0.84 & 2216.33 & 0.23 \\
Motion 3-to-4 & 0.18 & 0.74 & 1915.69 & 0.25 \\
ActionMesh  & 0.18 & \textbf{0.85} & 1410.61 & 0.22 \\
Ours        & \textbf{0.12} & 0.84 & \textbf{890.96} & \textbf{0.18} \\
\bottomrule
\end{tabular}
\caption{
\textbf{Quantitative evaluation on Consistent4D.}
\method{} achieves the best LPIPS, FVD, and DreamSim scores among feed-forward mesh animation baselines, indicating improved visual fidelity and temporal consistency on in-the-wild sequences.
}
\label{tab:consistent4d_quant}
\end{table}

\section{Implementation Details}
\label{sec:supp_implementation_details}
\paragraph{Preprocessing.}
We normalize each training sequence with respect to its first frame. The first-frame shape is centered and uniformly scaled so that the maximum side length of its bounding box is one, and the same transformation is applied to all subsequent frames. During training, we sample $10{,}000$ surface points from the canonical mesh and use their coordinates and surface normals as input to the positional encoding. We use the frozen Triposg shape encoder.
\paragraph{Architecture.}
Our model uses $31$ non-root latent motion components and one additional root component for global object motion. The hidden dimension is set to $512$ throughout the network. For visual conditioning, we extract frame-wise features from a frozen DINOv3 backbone at $512 \times 512$ resolution. The motion encoder contains $16$ blocks, each consisting of cross-attention to DINOv3 visual tokens, spatial self-attention across latent motion components within each frame, and temporal self-attention across frames for each latent component. We use an attention head dimension of $64$ and a dropout rate of $0.05$.

\paragraph{Decoders.}
Rigid transformations are predicted from the output motion latents using two-layer MLP decoders, with separate heads for rotation and translation. The softmax temperature for both skinning-weight prediction and the contrastive loss is set to $1$.

\paragraph{Training.}
We train the model with supervision on point offsets. For each frame, the ground-truth position of each sampled point is computed by applying its barycentric coordinates to the corresponding mesh vertices provided by the training data. We optimize the model using Adam with a learning rate of $2 \times 10^{-3}$ and apply gradient clipping. The model is trained for $1000$ epochs, taking approximately $1.5$ days on $8$ NVIDIA H200 GPUs. The TripoSG encoder, PartField encoder, and DINOv3 backbone are kept frozen throughout training, and only the newly introduced modules are optimized.

\paragraph{Inference.}
At inference time, our model processes $16$ frames at once. For longer videos, we apply the model autoregressively to obtain animations over the full sequence.
The factorized attention design also yields the fastest runtime among the compared methods, as shown in Table~\ref{tab:model_runtime_comparison}.

\begin{table}[t]
\centering

\begin{tabular}{lc}
\hline
\textbf{Model} & \textbf{Runtime} \\
\hline
ActionMesh & 350 s \\
Motion 3-to-4 & 4.50 s \\
Mesh4D & 55.33 s \\
Ours & \textbf{3.13} s \\
\hline
\end{tabular}
\caption{We report inference runtime across methods for 16 frame video on rtx 4080 }
\label{tab:model_runtime_comparison}
\end{table}

\begin{figure}[t]
    \centering
    \includegraphics[
        width=1.000\textwidth,
        height=1.000\textheight,
        keepaspectratio
    ]{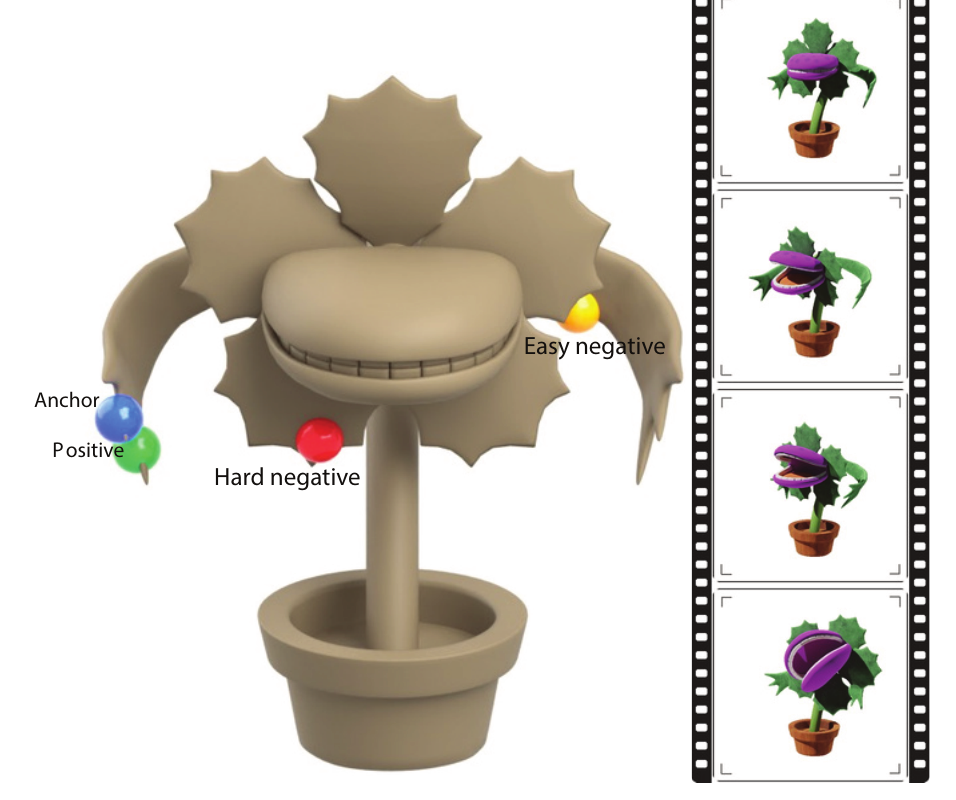}
\caption{
\textbf{Contrastive triplet construction.}
For each \textcolor{blue}{anchor} vertex, we sample \textcolor{green}{positive} vertices with similar motion and \textcolor{red}{hard negatives} from nearby vertices with dissimilar motion, along with \textcolor{orange}{easy negatives} from distant regions. This encourages motion-consistent vertices to learn similar skinning weights while separating vertices with different motion.
}
    \label{fig:contrastive_triplets}
\end{figure}

\section{Motion-aware Contrastive Loss}
\label{sec:supp_contrastive_loss}
We describe the implementation of the motion-aware contrastive loss used in
Sec.~\ref{sec:training}. The purpose of this loss is to encourage points with
similar ground-truth motion trajectories to have similar predicted non-root
skinning-weights.

\subsection{Trajectory Descriptor.}
For each point trajectory $\{\vx_{t,n}^{*}\}_{t=1}^{T}$, we construct a compact
motion descriptor from local temporal triplets
$(\vx_{t-1,n}^{*},\vx_{t,n}^{*},\vx_{t+1,n}^{*})$. Let
\begin{equation}
    \vu_{t,n}^{-}
    =
    \frac{\vx_{t,n}^{*}-\vx_{t-1,n}^{*}}
    {\|\vx_{t,n}^{*}-\vx_{t-1,n}^{*}\|_2+\epsilon},
    \qquad
    \vu_{t,n}^{+}
    =
    \frac{\vx_{t+1,n}^{*}-\vx_{t,n}^{*}}
    {\|\vx_{t+1,n}^{*}-\vx_{t,n}^{*}\|_2+\epsilon}.
\end{equation}
We define a rotation-axis direction and a translation direction as
\begin{equation}
    \vr_{t,n}
    =
    \frac{\vu_{t,n}^{-}\times \vu_{t,n}^{+}}
    {\|\vu_{t,n}^{-}\times \vu_{t,n}^{+}\|_2+\epsilon},
    \qquad
    \vh_{t,n}
    =
    \frac{\vu_{t,n}^{-}+\vu_{t,n}^{+}}
    {\|\vu_{t,n}^{-}+\vu_{t,n}^{+}\|_2+\epsilon}.
\end{equation}
To distinguish static points, we define
$b_{t,n}=I[s_{t,n}<\delta]$. The final per-frame motion descriptor is
\begin{equation}
    \vm_{t,n}
    =
    \left[
    (1-b_{t,n})\vr_{t,n},
    \;
    (1-b_{t,n})\vh_{t,n},
    \;
    b_{t,n}
    \right]
    \in \mathbb{R}^{7}.
\end{equation}
We normalize $\vm_{t,n}$ before computing pairwise similarities.

\subsection{Motion-based Pair Mining.}
For two points $n$ and $m$, we compute their trajectory similarity by averaging
cosine similarity over time:
\begin{equation}
    s^{\mathrm{mot}}_{n,m}
    =
    \frac{1}{T}
    \sum_{t=1}^{T}
    \left\langle
    \vm_{t,n},
    \vm_{t,m}
    \right\rangle .
\end{equation}
For an anchor point $a$, positives and negatives are selected using fixed motion
similarity thresholds:
\begin{equation}
    \mathcal{P}(a)
    =
    \{m \mid s^{\mathrm{mot}}_{a,m} > \theta_{+}\},
    \qquad
    \mathcal{N}(a)
    =
    \{m \mid s^{\mathrm{mot}}_{a,m} < \theta_{-}\}.
\end{equation}
We use $\theta_{+}=0.9$ and $\theta_{-}=0.5$. These values are a strong threshold for positives and negatives. 

Motion similarity determines the candidate positive and negative sets, while
canonical-space distance determines the sampling probabilities. Let
\begin{equation}
    d_{a,m}
    =
    \|\vx_{1,a}^{*}-\vx_{1,m}^{*}\|_2
\end{equation}
be the distance between points in the canonical frame. Positives are sampled
preferentially from nearby motion-consistent points:
\begin{equation}
    q^{+}_{a,m}
    \propto
    I[m\in\mathcal{P}(a)]
    \frac{1}{d_{a,m}}.
\end{equation}
Negatives are sampled from two groups. Easy negatives are spatially distant
points with dissimilar motion,
\begin{equation}
    q^{\mathrm{easy}}_{a,m}
    \propto
    I[m\in\mathcal{N}(a)]
    d_{a,m},
\end{equation}
while hard negatives are nearby points with dissimilar motion,
\begin{equation}
    q^{\mathrm{hard}}_{a,m}
    \propto
    I[m\in\mathcal{N}(a)]
    \frac{1}{d_{a,m}}.
\end{equation}
An example of this sampling is shown in Figure~\ref{fig:contrastive_triplets}. This sampling strategy encourages both global separation between unrelated
motions and sharp local transitions between neighboring parts that move
differently.

\subsection{Contrastive Objective.}
The contrastive loss is applied to the predicted non-root skinning weights
$\vw_n=\{w_{n,i}\}_{i=2}^{J}$. For two points $u$ and $v$, we measure similarity
between their predicted distributions using the Bhattacharyya coefficient:
\begin{equation}
    s^{\mathrm{skin}}_{u,v}
    =
    \sum_{i=2}^{J}
    \sqrt{w_{u,i}w_{v,i}} .
\end{equation}
Given an anchor $a$, a sampled positive $b\in\mathcal{P}(a)$, and sampled
negatives $\mathcal{N}_{s}(a)$, we use the symmetric InfoNCE loss
\begin{equation}
\begin{aligned}
\mathcal{L}_{\mathrm{con}}
= -\frac{1}{2}\Bigg[
&\log
\frac{
    \exp(s^{\mathrm{skin}}_{a,b}/\tau)
}{
    \exp(s^{\mathrm{skin}}_{a,b}/\tau)
    + \sum_{c\in\mathcal{N}_{s}(a)}
    \exp(s^{\mathrm{skin}}_{a,c}/\tau)
} \\
&+
\log
\frac{
    \exp(s^{\mathrm{skin}}_{b,a}/\tau)
}{
    \exp(s^{\mathrm{skin}}_{b,a}/\tau)
    + \sum_{c\in\mathcal{N}_{s}(b)}
    \exp(s^{\mathrm{skin}}_{b,c}/\tau)
}
\Bigg].
\end{aligned}
\end{equation}
where $\tau$ is the temperature. This objective encourages vertices with similar
trajectory descriptors to share similar latent skinning distributions, while
separating vertices that exhibit different motion.

\section{Broader Societal Impact.}
This work focuses on object-level 3D animation and can reduce the manual effort required to create animated 3D assets. Since the method operates on object geometry and motion, it does not directly involve sensitive personal data or decision-making systems. Potential misuse is mainly limited to creating misleading animated 3D content.

\clearpage
\section*{NeurIPS Paper Checklist}

\begin{enumerate}

\item {\bf Claims}
    \item[] Question: Do the main claims made in the abstract and introduction accurately reflect the paper's contributions and scope?
    \item[] Answer: \answerYes{} % Replace by \answerYes{}, \answerNo{}, or \answerNA{}.
    \item[] Justification: The abstract and introduction state the paper's main contributions, including the latent rigid motion representation, factorized motion-aware architecture, and accurate mesh animation objectives. The claims are supported by the method description and experimental evaluation.
    \item[] Guidelines:
    \begin{itemize}
        \item The answer \answerNA{} means that the abstract and introduction do not include the claims made in the paper.
        \item The abstract and/or introduction should clearly state the claims made, including the contributions made in the paper and important assumptions and limitations. A \answerNo{} or \answerNA{} answer to this question will not be perceived well by the reviewers. 
        \item The claims made should match theoretical and experimental results, and reflect how much the results can be expected to generalize to other settings. 
        \item It is fine to include aspirational goals as motivation as long as it is clear that these goals are not attained by the paper. 
    \end{itemize}

\item {\bf Limitations}
    \item[] Question: Does the paper discuss the limitations of the work performed by the authors?
    \item[] Answer: \answerYes{} % Replace by \answerYes{}, \answerNo{}, or \answerNA{}.
    \item[] Justification: The Discussion and Limitations section notes that suboptimal skinning weights can reduce animation quality, when visually the vertices are similar.  
    \item[] Guidelines:
    \begin{itemize}
        \item The answer \answerNA{} means that the paper has no limitation while the answer \answerNo{} means that the paper has limitations, but those are not discussed in the paper. 
        \item The authors are encouraged to create a separate ``Limitations'' section in their paper.
        \item The paper should point out any strong assumptions and how robust the results are to violations of these assumptions (e.g., independence assumptions, noiseless settings, model well-specification, asymptotic approximations only holding locally). The authors should reflect on how these assumptions might be violated in practice and what the implications would be.
        \item The authors should reflect on the scope of the claims made, e.g., if the approach was only tested on a few datasets or with a few runs. In general, empirical results often depend on implicit assumptions, which should be articulated.
        \item The authors should reflect on the factors that influence the performance of the approach. For example, a facial recognition algorithm may perform poorly when image resolution is low or images are taken in low lighting. Or a speech-to-text system might not be used reliably to provide closed captions for online lectures because it fails to handle technical jargon.
        \item The authors should discuss the computational efficiency of the proposed algorithms and how they scale with dataset size.
        \item If applicable, the authors should discuss possible limitations of their approach to address problems of privacy and fairness.
        \item While the authors might fear that complete honesty about limitations might be used by reviewers as grounds for rejection, a worse outcome might be that reviewers discover limitations that aren't acknowledged in the paper. The authors should use their best judgment and recognize that individual actions in favor of transparency play an important role in developing norms that preserve the integrity of the community. Reviewers will be specifically instructed to not penalize honesty concerning limitations.
    \end{itemize}

\item {\bf Theory assumptions and proofs}
    \item[] Question: For each theoretical result, does the paper provide the full set of assumptions and a complete (and correct) proof?
    \item[] Answer: \answerNA{} % Replace by \answerYes{}, \answerNo{}, or \answerNA{}.
    \item[] Justification: The paper does not present theoretical results, theorems, or formal proofs. The equations define the proposed representation, architecture, and training objectives.
    \item[] Guidelines:
    \begin{itemize}
        \item The answer \answerNA{} means that the paper does not include theoretical results. 
        \item All the theorems, formulas, and proofs in the paper should be numbered and cross-referenced.
        \item All assumptions should be clearly stated or referenced in the statement of any theorems.
        \item The proofs can either appear in the main paper or the supplemental material, but if they appear in the supplemental material, the authors are encouraged to provide a short proof sketch to provide intuition. 
        \item Inversely, any informal proof provided in the core of the paper should be complemented by formal proofs provided in appendix or supplemental material.
        \item Theorems and Lemmas that the proof relies upon should be properly referenced. 
    \end{itemize}

    \item {\bf Experimental result reproducibility}
    \item[] Question: Does the paper fully disclose all the information needed to reproduce the main experimental results of the paper to the extent that it affects the main claims and/or conclusions of the paper (regardless of whether the code and data are provided or not)?
    \item[] Answer: \answerYes{} % Replace by \answerYes{}, \answerNo{}, or \answerNA{}.
    \item[] Justification: The paper describes the model architecture, motion representation, training losses, datasets, baselines, and evaluation metrics needed to reproduce the main results. Additional implementation details are provided in the Appendix. 
    \item[] Guidelines:
    \begin{itemize}
        \item The answer \answerNA{} means that the paper does not include experiments.
        \item If the paper includes experiments, a \answerNo{} answer to this question will not be perceived well by the reviewers: Making the paper reproducible is important, regardless of whether the code and data are provided or not.
        \item If the contribution is a dataset and\slash or model, the authors should describe the steps taken to make their results reproducible or verifiable. 
        \item Depending on the contribution, reproducibility can be accomplished in various ways. For example, if the contribution is a novel architecture, describing the architecture fully might suffice, or if the contribution is a specific model and empirical evaluation, it may be necessary to either make it possible for others to replicate the model with the same dataset, or provide access to the model. In general. releasing code and data is often one good way to accomplish this, but reproducibility can also be provided via detailed instructions for how to replicate the results, access to a hosted model (e.g., in the case of a large language model), releasing of a model checkpoint, or other means that are appropriate to the research performed.
        \item While NeurIPS does not require releasing code, the conference does require all submissions to provide some reasonable avenue for reproducibility, which may depend on the nature of the contribution. For example
        \begin{enumerate}
            \item If the contribution is primarily a new algorithm, the paper should make it clear how to reproduce that algorithm.
            \item If the contribution is primarily a new model architecture, the paper should describe the architecture clearly and fully.
            \item If the contribution is a new model (e.g., a large language model), then there should either be a way to access this model for reproducing the results or a way to reproduce the model (e.g., with an open-source dataset or instructions for how to construct the dataset).
            \item We recognize that reproducibility may be tricky in some cases, in which case authors are welcome to describe the particular way they provide for reproducibility. In the case of closed-source models, it may be that access to the model is limited in some way (e.g., to registered users), but it should be possible for other researchers to have some path to reproducing or verifying the results.
        \end{enumerate}
    \end{itemize}

\item {\bf Open access to data and code}
    \item[] Question: Does the paper provide open access to the data and code, with sufficient instructions to faithfully reproduce the main experimental results, as described in supplemental material?
    \item[] Answer: \answerNo{} % Replace by \answerYes{}, \answerNo{}, or \answerNA{}.
    \item[] Justification: The source code is not provided at this time. The source code will be released on acceptance
    \item[] Guidelines:
    \begin{itemize}
        \item The answer \answerNA{} means that paper does not include experiments requiring code.
        \item Please see the NeurIPS code and data submission guidelines (\url{https://neurips.cc/public/guides/CodeSubmissionPolicy}) for more details.
        \item While we encourage the release of code and data, we understand that this might not be possible, so \answerNo{} is an acceptable answer. Papers cannot be rejected simply for not including code, unless this is central to the contribution (e.g., for a new open-source benchmark).
        \item The instructions should contain the exact command and environment needed to run to reproduce the results. See the NeurIPS code and data submission guidelines (\url{https://neurips.cc/public/guides/CodeSubmissionPolicy}) for more details.
        \item The authors should provide instructions on data access and preparation, including how to access the raw data, preprocessed data, intermediate data, and generated data, etc.
        \item The authors should provide scripts to reproduce all experimental results for the new proposed method and baselines. If only a subset of experiments are reproducible, they should state which ones are omitted from the script and why.
        \item At submission time, to preserve anonymity, the authors should release anonymized versions (if applicable).
        \item Providing as much information as possible in supplemental material (appended to the paper) is recommended, but including URLs to data and code is permitted.
    \end{itemize}

\item {\bf Experimental setting/details}
    \item[] Question: Does the paper specify all the training and test details (e.g., data splits, hyperparameters, how they were chosen, type of optimizer) necessary to understand the results?
    \item[] Answer: \answerYes{} % Replace by \answerYes{}, \answerNo{}, or \answerNA{}.
    \item[] Justification: All implementation details of the model,training and test details are provided in the main paper and the Appendix.
    \item[] Guidelines:
    \begin{itemize}
        \item The answer \answerNA{} means that the paper does not include experiments.
        \item The experimental setting should be presented in the core of the paper to a level of detail that is necessary to appreciate the results and make sense of them.
        \item The full details can be provided either with the code, in appendix, or as supplemental material.
    \end{itemize}

\item {\bf Experiment statistical significance}
    \item[] Question: Does the paper report error bars suitably and correctly defined or other appropriate information about the statistical significance of the experiments?
    \item[] Answer: \answerNo % Replace by \answerYes{}, \answerNo{}, or \answerNA{}.
    \item[] Justification:  Section 4 specifies the training dataset, evaluation datasets, baselines, metrics, and quantitative/qualitative evaluation. Note that we follow the evaluation protocol of prior feed-forward mesh animation work~\citep{sabathier2026actionmesh,jiang2026mesh4d,chen2026motion}, which reports single-run metrics on fixed benchmarks. Re-running with multiple random seeds at our scale(1.5 days x 8 H200 GPUs per run) is computationally prohibitive.   
    \item[] Guidelines:
    \begin{itemize}
        \item The answer \answerNA{} means that the paper does not include experiments.
        \item The authors should answer \answerYes{} if the results are accompanied by error bars, confidence intervals, or statistical significance tests, at least for the experiments that support the main claims of the paper.
        \item The factors of variability that the error bars are capturing should be clearly stated (for example, train/test split, initialization, random drawing of some parameter, or overall run with given experimental conditions).
        \item The method for calculating the error bars should be explained (closed form formula, call to a library function, bootstrap, etc.)
        \item The assumptions made should be given (e.g., Normally distributed errors).
        \item It should be clear whether the error bar is the standard deviation or the standard error of the mean.
        \item It is OK to report 1-sigma error bars, but one should state it. The authors should preferably report a 2-sigma error bar than state that they have a 96\% CI, if the hypothesis of Normality of errors is not verified.
        \item For asymmetric distributions, the authors should be careful not to show in tables or figures symmetric error bars that would yield results that are out of range (e.g., negative error rates).
        \item If error bars are reported in tables or plots, the authors should explain in the text how they were calculated and reference the corresponding figures or tables in the text.
    \end{itemize}

\item {\bf Experiments compute resources}
    \item[] Question: For each experiment, does the paper provide sufficient information on the computer resources (type of compute workers, memory, time of execution) needed to reproduce the experiments?
    \item[] Answer: \answerYes{} % Replace by \answerYes{}, \answerNo{}, or \answerNA{}.
    \item[] Justification: Compute resources are mentioned in the Appendix.
    \item[] Guidelines:
    \begin{itemize}
        \item The answer \answerNA{} means that the paper does not include experiments.
        \item The paper should indicate the type of compute workers CPU or GPU, internal cluster, or cloud provider, including relevant memory and storage.
        \item The paper should provide the amount of compute required for each of the individual experimental runs as well as estimate the total compute. 
        \item The paper should disclose whether the full research project required more compute than the experiments reported in the paper (e.g., preliminary or failed experiments that didn't make it into the paper). 
    \end{itemize}
    
\item {\bf Code of ethics}
    \item[] Question: Does the research conducted in the paper conform, in every respect, with the NeurIPS Code of Ethics \url{https://neurips.cc/public/EthicsGuidelines}?
    \item[] Answer: \answerYes{} % Replace by \answerYes{}, \answerNo{}, or \answerNA{}.
    \item[] Justification: The work uses existing 3D animation datasets and does not involve human subjects, private data, or deployed decision-making systems. To the best of our knowledge, the research conforms to the NeurIPS Code of Ethics.
    \item[] Guidelines:
    \begin{itemize}
        \item The answer \answerNA{} means that the authors have not reviewed the NeurIPS Code of Ethics.
        \item If the authors answer \answerNo, they should explain the special circumstances that require a deviation from the Code of Ethics.
        \item The authors should make sure to preserve anonymity (e.g., if there is a special consideration due to laws or regulations in their jurisdiction).
    \end{itemize}

\item {\bf Broader impacts}
    \item[] Question: Does the paper discuss both potential positive societal impacts and negative societal impacts of the work performed?
    \item[] Answer: \answerYes{} % Replace by \answerYes{}, \answerNo{}, or \answerNA{}.
    \item[] Justification: The work may positively impact animation and 3D content creation by reducing manual rigging effort. Potential negative impacts include misuse for generating misleading animated 3D content. Mentioned in Appendix.
    \item[] Guidelines:
    \begin{itemize}
        \item The answer \answerNA{} means that there is no societal impact of the work performed.
        \item If the authors answer \answerNA{} or \answerNo, they should explain why their work has no societal impact or why the paper does not address societal impact.
        \item Examples of negative societal impacts include potential malicious or unintended uses (e.g., disinformation, generating fake profiles, surveillance), fairness considerations (e.g., deployment of technologies that could make decisions that unfairly impact specific groups), privacy considerations, and security considerations.
        \item The conference expects that many papers will be foundational research and not tied to particular applications, let alone deployments. However, if there is a direct path to any negative applications, the authors should point it out. For example, it is legitimate to point out that an improvement in the quality of generative models could be used to generate Deepfakes for disinformation. On the other hand, it is not needed to point out that a generic algorithm for optimizing neural networks could enable people to train models that generate Deepfakes faster.
        \item The authors should consider possible harms that could arise when the technology is being used as intended and functioning correctly, harms that could arise when the technology is being used as intended but gives incorrect results, and harms following from (intentional or unintentional) misuse of the technology.
        \item If there are negative societal impacts, the authors could also discuss possible mitigation strategies (e.g., gated release of models, providing defenses in addition to attacks, mechanisms for monitoring misuse, mechanisms to monitor how a system learns from feedback over time, improving the efficiency and accessibility of ML).
    \end{itemize}
    
\item {\bf Safeguards}
    \item[] Question: Does the paper describe safeguards that have been put in place for responsible release of data or models that have a high risk for misuse (e.g., pre-trained language models, image generators, or scraped datasets)?
    \item[] Answer: \answerNA{} % Replace by \answerYes{}, \answerNo{}, or \answerNA{}.
    \item[] Justification: The paper does not release a high-risk foundation model or scraped dataset.
    \item[] Guidelines:
    \begin{itemize}
        \item The answer \answerNA{} means that the paper poses no such risks.
        \item Released models that have a high risk for misuse or dual-use should be released with necessary safeguards to allow for controlled use of the model, for example by requiring that users adhere to usage guidelines or restrictions to access the model or implementing safety filters. 
        \item Datasets that have been scraped from the Internet could pose safety risks. The authors should describe how they avoided releasing unsafe images.
        \item We recognize that providing effective safeguards is challenging, and many papers do not require this, but we encourage authors to take this into account and make a best faith effort.
    \end{itemize}

\item {\bf Licenses for existing assets}
    \item[] Question: Are the creators or original owners of assets (e.g., code, data, models), used in the paper, properly credited and are the license and terms of use explicitly mentioned and properly respected?
    \item[] Answer: \answerYes{} % Replace by \answerYes{}, \answerNo{}, or \answerNA{}.
    \item[] Justification:  The paper cites the existing datasets, baselines, and pretrained models used in the experiments.
    \item[] Guidelines:
    \begin{itemize}
        \item The answer \answerNA{} means that the paper does not use existing assets.
        \item The authors should cite the original paper that produced the code package or dataset.
        \item The authors should state which version of the asset is used and, if possible, include a URL.
        \item The name of the license (e.g., CC-BY 4.0) should be included for each asset.
        \item For scraped data from a particular source (e.g., website), the copyright and terms of service of that source should be provided.
        \item If assets are released, the license, copyright information, and terms of use in the package should be provided. For popular datasets, \url{paperswithcode.com/datasets} has curated licenses for some datasets. Their licensing guide can help determine the license of a dataset.
        \item For existing datasets that are re-packaged, both the original license and the license of the derived asset (if it has changed) should be provided.
        \item If this information is not available online, the authors are encouraged to reach out to the asset's creators.
    \end{itemize}

\item {\bf New assets}
    \item[] Question: Are new assets introduced in the paper well documented and is the documentation provided alongside the assets?
    \item[] Answer: \answerNA{} % Replace by \answerYes{}, \answerNo{}, or \answerNA{}.
    \item[] Justification: No new asset is used in the paper
    \item[] Guidelines:
    \begin{itemize}
        \item The answer \answerNA{} means that the paper does not release new assets.
        \item Researchers should communicate the details of the dataset\slash code\slash model as part of their submissions via structured templates. This includes details about training, license, limitations, etc. 
        \item The paper should discuss whether and how consent was obtained from people whose asset is used.
        \item At submission time, remember to anonymize your assets (if applicable). You can either create an anonymized URL or include an anonymized zip file.
    \end{itemize}

\item {\bf Crowdsourcing and research with human subjects}
    \item[] Question: For crowdsourcing experiments and research with human subjects, does the paper include the full text of instructions given to participants and screenshots, if applicable, as well as details about compensation (if any)? 
    \item[] Answer: \answerNA{} % Replace by \answerYes{}, \answerNo{}, or \answerNA{}.
    \item[] Justification: No crowdsourcing experiment or human subject experiment are performed in the paper.
    \item[] Guidelines:
    \begin{itemize}
        \item The answer \answerNA{} means that the paper does not involve crowdsourcing nor research with human subjects.
        \item Including this information in the supplemental material is fine, but if the main contribution of the paper involves human subjects, then as much detail as possible should be included in the main paper. 
        \item According to the NeurIPS Code of Ethics, workers involved in data collection, curation, or other labor should be paid at least the minimum wage in the country of the data collector. 
    \end{itemize}

\item {\bf Institutional review board (IRB) approvals or equivalent for research with human subjects}
    \item[] Question: Does the paper describe potential risks incurred by study participants, whether such risks were disclosed to the subjects, and whether Institutional Review Board (IRB) approvals (or an equivalent approval/review based on the requirements of your country or institution) were obtained?
    \item[] Answer: \answerNA{} % Replace by \answerYes{}, \answerNo{}, or \answerNA{}.
    \item[] Justification: The paper does not involve human subjects research, so IRB approval or equivalent review is not applicable.
    \item[] Guidelines:
    \begin{itemize}
        \item The answer \answerNA{} means that the paper does not involve crowdsourcing nor research with human subjects.
        \item Depending on the country in which research is conducted, IRB approval (or equivalent) may be required for any human subjects research. If you obtained IRB approval, you should clearly state this in the paper. 
        \item We recognize that the procedures for this may vary significantly between institutions and locations, and we expect authors to adhere to the NeurIPS Code of Ethics and the guidelines for their institution. 
        \item For initial submissions, do not include any information that would break anonymity (if applicable), such as the institution conducting the review.
    \end{itemize}

\item {\bf Declaration of LLM usage}
    \item[] Question: Does the paper describe the usage of LLMs if it is an important, original, or non-standard component of the core methods in this research? Note that if the LLM is used only for writing, editing, or formatting purposes and does \emph{not} impact the core methodology, scientific rigor, or originality of the research, declaration is not required.
    %this research? 
    \item[] Answer: \answerNA{} % Replace by \answerYes{}, \answerNo{}, or \answerNA{}.
    \item[] Justification: The core method development does not use LLMs as an important, original, or non-standard component.
    \item[] Guidelines:
    \begin{itemize}
        \item The answer \answerNA{} means that the core method development in this research does not involve LLMs as any important, original, or non-standard components.
        \item Please refer to our LLM policy in the NeurIPS handbook for what should or should not be described.
    \end{itemize}

\end{enumerate}
\end{document}